%% file: ms.tex
\documentclass{bmvc2k}

\title{MoManifold: Learning to Measure 3D Human Motion via Decoupled Joint Acceleration Manifolds}

\addauthor{Ziqiang Dang}{ZiqDang@zju.edu.cn}{1*}
\addauthor{Tianxing Fan}{tianxingfan@zju.edu.cn}{1*}
\addauthor{Boming Zhao}{bmzhao@zju.edu.cn}{1}
\addauthor{Xujie Shen}{shenfishcrap@gmail.com}{1}
\addauthor{Lei Wang}{wanglei12@oppo.com}{2}
\addauthor{Guofeng Zhang}{zhangguofeng@zju.edu.cn}{1}
\addauthor{Zhaopeng Cui}{zhpcui@zju.edu.cn}{1†}

\addinstitution{
 State Key Lab of CAD\&CG,\\
 Zhejiang University,\\
 Hangzhou, China
}
\addinstitution{
 Guangdong OPPO Mobile Telecommunications Corp Ltd,\\
 Guangdong, China
}

\runninghead{DANG ET AL.}{MoManifold}

\def\eg{\emph{e.g}\bmvaOneDot}

\def\etal{\emph{et al}\bmvaOneDot}
\def\ie{\emph{i.e}\bmvaOneDot}

\usepackage{color}
\definecolor{myred}{rgb}{1.0,0.0,0.0}
\newcommand{\zhaopeng}[1]{\textcolor{myred}{#1}}
\definecolor{myblue}{rgb}{0.0,0.0,0.0}
\newcommand{\ziqiang}[1]{\textcolor{myblue}{#1}}

\newcommand{\ssecspace}{\vspace{-0.5em}}
\newcommand{\boldparagraph}[1]{\noindent{\bf #1} }

\usepackage{adjustbox}
\usepackage{bibentry}
\usepackage{xspace}
\usepackage{amsmath}
\usepackage{amssymb}
\usepackage{array}
\usepackage{tabularray}
\usepackage{multirow}
\usepackage{perpage}
\MakePerPage{footnote}
\input{Sections/var}
\begin{document}
\maketitle
\input{Sections/00_abstract}
\input{Sections/01_intro}
\input{Sections/02_related}
\input{Sections/03_method}
\input{Sections/04_exp}
\input{Sections/05_conclusion}
\bibliography{egbib}

\end{document}


\maketitle
\appendix
\input{Sections/Outline}
\input{Sections/A_implementation_details}
\input{Sections/B_Experimental_Evaluation_Details}
\input{Sections/C_Extended_Experiments}
\input{Sections/D_Ablation_Study}

\input{Sections/E_Discussions}
\clearpage
\bibliography{supp}

%% file: Sections/var.tex
\def\RthreeT{$\mathbb{R}^{\left(T-2\right) \times 3}$}

\def\method{MoManifold\xspace}

\def\accel{$\boldsymbol{\vec{\alpha}}=\left \{ \left ( \alpha_{x_1},\alpha_{y_1},\alpha_{z_1} \right ) ,\dots,\left ( \alpha_{x_{T-2}},\alpha_{y_{T-2}},\alpha_{z_{T-2}} \right ) \right \}$}

\def\accelhata{$\boldsymbol{\hat{\vec{\alpha}}}=\{ (\hat{\alpha}_{x_1},\hat{\alpha}_{y_1},\hat{\alpha}_{z_1} ) ,\dots,( \hat{\alpha}_{x_{T-2}},$}
\def\accelhatb{$\hat{\alpha}_{y_{T-2}},\hat{\alpha}_{z_{T-2}}) \}$}

\def\alphavec{$\boldsymbol{\vec{\alpha}}$}
\def\alphavechat{$\boldsymbol{\hat{\vec{\alpha}}}$}

%% file: Sections/00_abstract.tex
\begin{abstract}
Incorporating temporal information effectively is important for accurate 3D human motion estimation and generation which have wide applications from human-computer interaction to AR/VR. In this paper, we present \method, a novel human motion prior, which models plausible human motion in continuous high-dimensional motion space. Different from existing mathematical or VAE-based methods, our representation is designed based on the neural distance field, which makes human dynamics explicitly quantified to a score and thus can measure human motion plausibility. Specifically, we propose novel decoupled joint acceleration manifolds to model human dynamics from existing limited motion data. 
Moreover, we introduce a novel optimization method using the manifold distance as guidance, which facilitates a variety of motion-related tasks. 
Extensive experiments demonstrate that \method outperforms existing SOTAs as a prior in several downstream tasks such as denoising real-world human mocap data, recovering human motion from partial 3D observations, mitigating jitters for SMPL-based pose estimators, and refining the results of motion in-betweening.

\end{abstract}

%% file: Sections/01_intro.tex
\section{Introduction}
\label{sec:intro}

3D human motion estimation aims to predict the 3D spatial configurations and trajectories of the human body over time, and it is essential for human behaviour understanding with wide applications from surveillance and human-computer interaction to virtual reality and augmented reality. 
Although extensive efforts have been proposed for 3D pose and shape estimation from a single image~\cite{Kocabas_PARE_2021,pymaf2021}, 
these methods inevitably lead to jitter artifacts or unnatural motions due to self-occlusion and partial observations, which cannot be easily addressed by simple pose optimization with additional temporal regularization terms (\eg, the sum of joint differences or mesh vertex differences between consecutive frames) \cite{zeng2022smoothnet,zhang2021learning}, since such optimization terms will enforce the differences between consecutive frames to be zero, leading the optimization process towards static motion, hindering the natural motion dynamics.


To improve the performance of 3D human motion estimation, many approaches are proposed to incorporate human motion priors. Pioneer works exploit mathematical models like PCA~\cite{ormoneit2000learning} and GDPMs~\cite{urtasun20063d} to learn the temporal motion priors, while these methods are limited to simple or specific motion. With the development of deep learning, several recurrent and autoregressive models~\cite{henter2020moglow,yang2021lobstr} are proposed to learn the sequential nature of human motion. However, these methods normally suffer from careful tuning to handle run-time user requests and error accumulation for long sequences. Recently, the VAE-based methods~\cite{ling2020character,rempe2021humor} are proposed to learn the plausible motion space, while these methods tend to produce average motion by folding a manifold into a Gaussian distribution~\cite{tiwari22posendf}.

In this paper, we present a novel human motion prior, \ie, \method, which models the plausible human motion in a continuous high-dimensional motion space. 
Compared to existing pose priors \cite{SMPL-X:2019,tiwari22posendf} which model the human pose of a single frame,  modeling human motion is more challenging. Pose priors focus on static poses, while our motion prior aims to address the dynamics of continuous human movements.
Besides, different from existing motion priors, our representation is designed based on the neural distance field, which allows for explicit quantification of human dynamics, providing a distance score to measure the motion plausibility. A larger distance represents a departure from the manifold of natural human motion, indicating potential anomalies in motion. Benefiting from such modeling, our MoManifold empowers a variety of tasks, such as denoising real-world human mocap data, recovering human motion from partial 3D observations, jitter mitigation for human pose estimators, and refining the results of motion in-betweening. 

However, it is nontrivial to design such a representation.
At first, different from the single-frame pose, the human motion encapsulates sequences of poses over time and inherently increases the dimensionality of the data. Thus it is hard to learn to map the naively concatenated poses at different timesteps to a distance value because a dramatic increase of training data is inevitable while impossible to fulfill given the existing human motion datasets.
To handle this problem, we propose to learn the manifold of plausible acceleration vectors for each body joint individually in high dimensional space, where an acceleration vector is defined as a point represented by the acceleration of T frames' motion. The distance to the manifold measures whether the joint motion complies with human dynamics. 
By decoupling the joints, we substantially reduce the input dimension, thus ensuring the successful learning of implicit surfaces from existing limited motion data (\eg, from an input dimension of 1008 to 42 when considering 16 frames). Despite the decoupling, these joints maintain an inherent correlation through the SMPL model topology and thus reflect human dynamics as a whole. In other words, as long as each joint’s movement is plausible, the human motion is feasible. 
Moreover, different body parts have specific motion characteristics, and we cannot naively combine different joint acceleration manifolds. 
As a result, we adopt 
a weighted design based on human skeleton geometry to better model human dynamics. 
At last, for downstream tasks, we introduce a novel optimization method based on \method, which utilizes the distance as guidance for optimization and integrates with a traditional temporal regularization term based on the characteristics of different joints to help jump out of local optima.

Our contributions are summarized as follows: 1) We present a novel human motion prior, \ie, \method, which models plausible human motion in a continuous high-dimensional motion space and can be used to measure human motion plausibility, thus facilitating downstream tasks such as denoising real-world human mocap data, recovering human motion from partial 3D observations, jitter mitigation for human pose estimators and refining the results of motion in-betweening. 2) Decoupled joint acceleration manifolds and a weighted design based on human skeleton geometry are adopted to model human dynamics to deal with the dramatic demand for human motion training data. 3) We introduce a novel motion optimization method based on \method, which can be applied to various downstream tasks. 
4) Extensive experiments demonstrate that \method has good generalization ability and outperforms existing SOTAs on multiple motion-related tasks.

%% file: Sections/02_related.tex
\section{Related Work}
\label{sec:related-work}
\boldparagraph{Pose and Motion Priors.}
Human pose and motion priors play a crucial role in human-centered research and applications, guiding to produce more accurate and realistic human poses and movements.
Regarding pose priors, early research primarily concentrated on learning constraints for joint limits~\cite{shao2003general,engell2012joint,akhter2015pose}. 
SMPLify~\cite{Bogo:ECCV:2016} fits a Gaussian Mixture Model (GMM) to a motion capture dataset and uses the GMM for downstream tasks~\cite{alldieck19cvpr,bhatnagar2019mgn,tiwari20sizer} to preserve the realism of poses. Recently, some studies have utilized deep learning methods to learn pose priors. VPoser~\cite{SMPL-X:2019} learns a compact representation space to constrain the poses. HMR~\cite{hmrKanazawa17} and VIBE~\cite{kocabas2019vibe} learn the pose prior by adversarial loss in the training process of their own tasks. Pose-NDF~\cite{tiwari22posendf} learns a continuous model for plausible human poses. GAN-S prior~\cite{Davydov_2022_CVPR} introduces GAN-based pose prior and outperforms the VAE-based prior. 
Regarding motion priors, VIBE~\cite{kocabas2019vibe} learns the motion prior by adversarial loss. MotionVAE~\cite{ling2020character} employs autoregressive CVAE to learn distribution of the change in poses. HuMoR~\cite{rempe2021humor} is similar to MotionVAE, but generalizes to unseen, non-periodic motions. \ziqiang{
Recently, NeMF~\cite{he2022nemf} also designed a VAE-based motion prior, generating motion by sampling from the latent space, primarily for motion generation and editing applications. However, these VAE-based methods encode motion into a latent code $z$, which does not allow for an explicit measurement of motion plausibility. In addition, there have been recent works that establish a connection between motion and text~\cite{petrovich22temos,tevet2023human,chen2023executing,Lin_2023_CVPR}. It is worth noting that diffusion models~\cite{ho2020denoising,song2020denoising} have recently been applied to human motion modeling~\cite{chen2023executing,tevet2023human} and have achieved state-of-the-art results in text-guided motion generation task.}


\boldparagraph{3D Human Pose and Shape Estimation.}
The existing estimation methods can be divided into two categories, depending on whether they are optimization- or regression-based. The optimization-based methods directly optimize to more accurately fit to observations \eg, images or 2D/3D joint locations. SMPLify~\cite{Bogo:ECCV:2016} was the first method to fit the SMPL model to the output of a 2D keypoints detector. 
For motion sequences, several works~\cite{arnab2019exploiting,xiang2019monocular,liu20214d} apply simple smoothness optimization term over time. 
The regression-based methods directly regress the SMPL parameters from pixels of an input image~\cite{hmrKanazawa17,Kocabas_PARE_2021,pymafx2023,lin2023one} or video~\cite{kocabas2019vibe,wan2021,WeiLin2022mpsnet,shen2023global,goel2023humans,ye2023slahmr}. \ziqiang{Whether applying image-based methods directly to videos or using video-based methods that model temporal constraints, these approaches often suffer from severe jitters caused by rarely seen or occluded actions.}

\boldparagraph{Human Motion Smoothing.}
Existing learning-based motion smooth strategies can be classified into two types: Strategies embedded in its own models and refinement networks after estimators. For the former category, these methods apply various temporal architectures (\eg, GRUs~\cite{choi2021beyond}, 
Transformers~\cite{WeiLin2022mpsnet}) for temporal feature extraction to ensure smooth motion. 
For refinement networks, Jiang \etal~\cite{jiang2021skeletor} designed a transformer-based network to smooth 3D poses. Zeng \etal~\cite{zeng2022smoothnet} proposed SmoothNet to model the natural smoothness characteristics in body movements.

%% file: Sections/03_method.tex
\section{Method}
\label{sec:method}
We introduce \method, a 3D human motion prior which can well preserve human motion dynamics. We model the manifold of plausible human motion as the neural distance field which makes human dynamics explicitly quantified to a score (\ie, distance).

\subsection{Decoupled Joint Acceleration Manifolds}
\paragraph{Preliminaries: Body Model.}
The SMPL model~\cite{SMPL:2015} is a differentiable function that outputs a posed 3D human mesh $\mathcal{M}\left(\theta,\beta\right)\in\mathbb{R}^{6890\times3}$, given the pose parameter $\theta \in \mathbb{R}^{72}$ and shape parameter $\beta \in \mathbb{R}^{10}$. 
In this work, we leave the SMPL shape parameters $\beta$ untouched as in previous works~\cite{SMPL-X:2019,tiwari22posendf,Davydov_2022_CVPR}. The 3D joint locations $\mathcal{J}_{3D} = W \mathcal{M} \in \mathbb{R}^{K\times3}, K=24$, are computed with a pretrained linear regressor $\mathit{W}$. And we use the obtained 3D joints to calculate the acceleration vectors.

\begin{figure*}[!t]
    \addtolength{\belowcaptionskip}{-0.3cm}
    \centering
    \includegraphics[width=0.95\linewidth, trim={0 0 0 1mm}, clip]{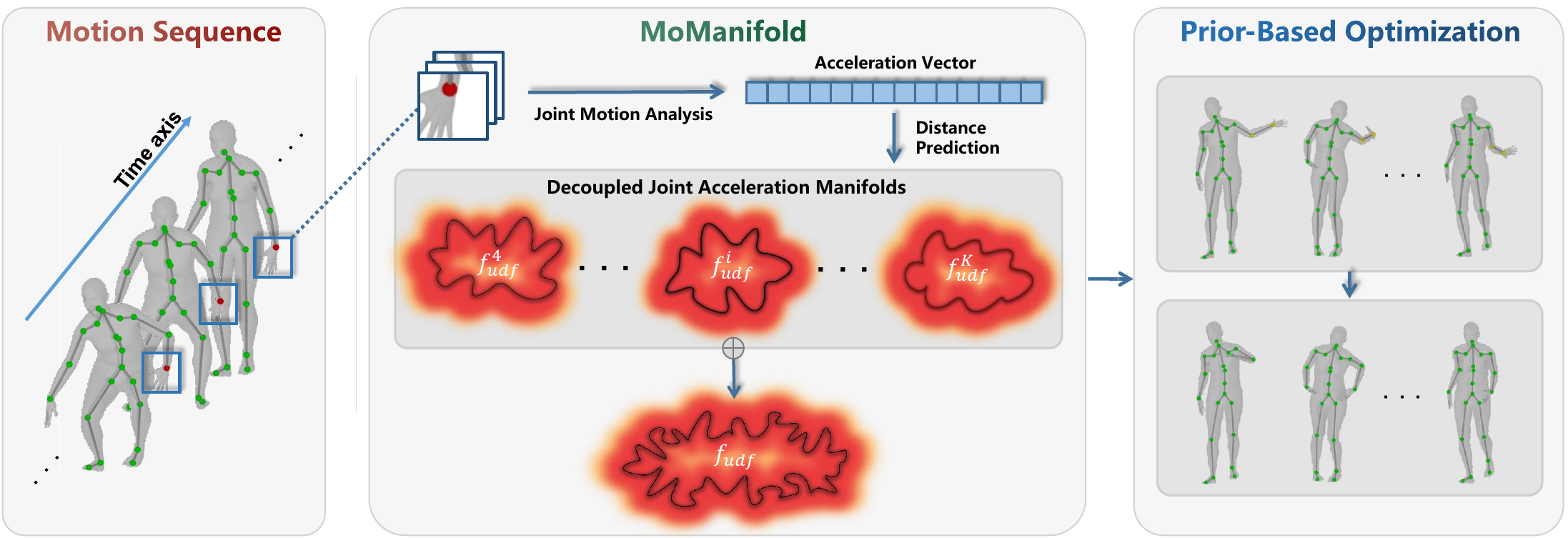}
    \vspace{-0.6em}
    \caption{
    \textbf{Overview of \method.} A motion sequence can be divided into different motion segments, represented as displacement vectors of body joints. Instead of directly learning the implicit surface of motion segment $\mathbf{m}$, \method learns an independent implicit surface of plausible acceleration vectors for every joint, and the distance to the manifold measures whether the joint motion complies with human dynamics. With a weighted design based on skeleton, we combine these manifolds to obtain the manifold of motion segments.
    }
    \label{fig:framework}
\end{figure*}

 As shown in Fig.~\ref{fig:framework}, we consider that a human motion sequence is composed of multiple short motion segments of T frames. A short motion segment can be represented as displacement vectors of human body joints, $\mathbf{m} \in \mathbb{R}^{T \times K \times 3}$. 
 We propose decoupled joint acceleration manifolds represented as unsigned distance fields (udf) for the modeling of plausible motion manifold.
 Instead of directly learning the implicit surface of motion segment $\mathbf{m}$, \method learns an independent implicit surface of plausible acceleration vectors for every body joint in high dimensional space \RthreeT, where an acceleration vector is defined as a point, represented by the acceleration of T frames' motion, and the distance to the manifold measures whether the joint motion complies with human dynamics. Then, we combine the learned implicit surfaces to construct the unsigned distance field of motion segment $\mathbf{m}$. 

 Given a neural network $f^{i}_{udf} : \mathbb{R}^{\left(T-2\right) \times 3} \longmapsto \mathbb{R}^{+}$, which maps an acceleration vector of the joint $i$, \alphavec \space$ \in \mathbb{R}^{(T-2) \times 3}$, to a non-negative scalar, we formulate the manifold of plausible acceleration vectors as the zero level set:
\begin{equation}
  S=\left \{ \boldsymbol{\vec{\alpha}} \in \mathbb{R}^{(T-2) \times 3} |f^{i}_{udf}\left (  \boldsymbol{\vec{\alpha}}\right )=0 \right \}.
  \label{eq:zero-level}
\end{equation}

Thus, we can obtain K unsigned distance fields for the K joints. 
We empirically find that 
it is difficult to learn implicit surfaces for joints \{$J_{0}$,$J_{1}$,$J_{2}$,$J_{3}$\}, 
because in the relative coordinate system, the positions of these joints remain largely static, their acceleration vectors undergo extremely subtle and limited variations, making them difficult to be captured. Thus, we excluded the distance fields of these four joints. \ziqiang{For more details about these four joints, please refer to our supplementary material.}

We define the distance  $d:\mathbb{R}^{(T-2) \times 3} \times \mathbb{R}^{(T-2) \times 3} \to \mathbb{R}^{+}$ between two acceleration vectors \alphavec \space and \alphavechat \space as:
\begin{equation}
\medmuskip=0.5mu 
  d\left ( \boldsymbol{\vec{\alpha}},\boldsymbol{\hat{\vec{\alpha}}} \right )=\sum_{i=1}^{T-2}\left | \alpha_{x_i}-\hat{\alpha}_{x_i} \right |+\left | \alpha_{y_i}-\hat{\alpha}_{y_i} \right |+\left | \alpha_{z_i}-\hat{\alpha}_{z_i} \right |,
\label{eq:distance}
\end{equation}
where \accel, and \accelhata \\ \accelhatb \space represent the acceleration of T frames’ joint motion.

We use simple yet effective fully-connected networks to fit the unsigned distance fields of body joints. To construct the unsigned distance field for motion segment $\mathbf{m}$, we propose a compositional implicit neural function $f_{udf}$, which takes $\mathbf{\ddot{m}}=\left \{ \boldsymbol{\vec{\alpha}_{4}},\dots ,\boldsymbol{\vec{\alpha}_{K}} \right \} $, the acceleration of the motion segment $\mathbf{m}$ as input:
\vspace{-0.3cm}
\begin{equation}
  f_{udf}(\mathbf{\ddot{m}})=\sum_{i=4}^{K}w_{i}f^{i}_{udf}(\boldsymbol{\vec{\alpha}_{i}}),
\label{eq:m_field}
\vspace{-0.2cm}
\end{equation}
where $w_{i}$ is the weight associated with each joint 
and determined by the summation of bone lengths from the corresponding joint to the root joint along the kinematic structure of the SMPL body model (\ie, later joints in the chain have larger weights), and $f^{i}_{udf}$ is the implicit neural function of the joint~$i$ that predicts the unsigned distance for the given acceleration vector $\boldsymbol{\vec{\alpha}_{i}}$.
For detailed description of the weighted design, please refer to our supplementary.

\subsection{Data Preparation and Training Loss}
\label{subsec:loss}
To train unsigned distance fields, we randomly sample motion segments from AMASS \cite{AMASS:ICCV:2019}, a comprehensive motion capture database, and consider these as zero-level (distance = 0). Additionally, to obtain data with non-zero distances, we use artificially noised motion from AMASS as well as the estimated results from a representative human pose estimator VIBE \cite{kocabas2019vibe} on the MPI-INF-3DHP dataset~\cite{mono-3dhp2017}. 
For each acceleration vector, we identify the top-k nearest neighbors in zero-level and compute the average distance of Eq.~\eqref{eq:distance} as its distance value, where Faiss \cite{johnson2019billion} is utilized for efficient similarity search in dense vectors. \ziqiang{For more details about data preparation, please refer to our supplementary material.}

Due to different motion characteristics, for each body joint, its unsigned distance field $f^{i}_{udf}$ is trained independently using data pairs $\left ( \vec{\alpha},d \right )$. Instead of mapping to the geodesic distances, $f^{i}_{udf}$ learns a distance variant  as the following:
\vspace{-0.2cm}
\begin{equation}
  \mathcal{L}_{udf}=\left \| f^{i}_{udf}\left ( \vec{\alpha} \right )-\ln_{}{\left ( d+1 \right ) }   \right \|^{2}.
\label{eq:d_variant}
\vspace{-0.1cm}
\end{equation}

This loss function favors accurate distance prediction for points nearer to the manifold by logarithmically scaling distances, effectively regularizing the model to focus on points close to the manifold and diminish the influence of distant points. 
Additionally, we utilized the Eikonal regularizer $\mathcal{L}_{eikonal}$, which encourages a unit-norm gradient for the distance field outside the manifold~\cite{tiwari22posendf,icml2020_2086}:
\begin{equation}
  \mathcal{L}_{eikonal}=\left (  \left \|  \nabla_{\vec{\alpha}}f^{i}_{udf}\left ( \vec{\alpha} \right )  \right\|_{2}-1\right )^2.
\label{eq:eikonal}
\end{equation}

Thus, our final loss for each implicit neural function is then defined as:
\vspace{-0.1cm}
\begin{equation}
  \mathcal{L}_{total}=\lambda _{1}\mathcal{L}_{udf}+\lambda _{2}\mathcal{L}_{eikonal},
\label{eq:loss}
\vspace{-0.1cm}
\end{equation}
where $\lambda _{1}$ and $\lambda _{2}$ are loss weights.

\subsection{MoManifold As a Motion Prior}

After modeling human motion as an unsigned distance field, we can utilize \method as a motion prior for downstream tasks. Here, we introduce a novel optimization method that employs the distance value as a guiding metric for optimization and integrates with a traditional temporal term.

In optimization-based tasks, traditional temporal regularization terms (\eg, the sum of joint differences between consecutive frames) are normally employed to constrain the motion to be smooth enough. However, such optimization terms enforce the differences between consecutive frames to be zero, leading the optimization process towards static motion, hindering the natural motion dynamics. At the same time, they also compete with other optimization objectives, constraining their optimization directions, when the motion has already been well-regularized by temporal terms. The most commonly used temporal term is,
\vspace{-0.1cm}
\begin{equation}
  \varepsilon_{temp}=\sum_{t=1}^{T}\sum_{i=1}^{K}\left \| \mathbf{p}_{t}^{i}-\mathbf{p}_{t-1}^{i} \right \| _2,
\label{eq:tradition}
\vspace{-0.1cm}
\end{equation}
where $\mathbf{p}_{t}^{i} \in \mathbf{J}_{t}^{K}$ is the SMPL body joint $i$ of frame $t$.

Because \method explicitly quantifies human dynamics to a distance value, it can be used as a temporal regularization term to regularize the pose parameters of SMPL model: 
\begin{equation}
  \varepsilon_{motion}= f_{udf}(\mathbf{\ddot{m}}).
\label{eq:mo_term}
\end{equation}
Through learning the implicit surface of plausible motion, \method makes the optimization direction no longer the static motion, but the plausible motion of the manifold.

Moreover, in our supplementary material, we demonstrate that when fused with a traditional temporal regularization term Eq.~\eqref{eq:tradition}, \method can help achieve better optimization results. 
Thus, the fusion term for the optimization task is:
\begin{equation}
\medmuskip=1mu 
  \varepsilon_{fusion}= \varepsilon_{motion}+\sum_{i=1}^{K}\sum_{t=1}^{T}\left ( 1-w_{i} \right ) \left \| \mathbf{p}_{t}^{i}-\mathbf{p}_{t-1}^{i} \right \| _2,
\label{eq:fusion_term}
\end{equation}
where $w_{i}$ corresponds to $w_{i}$ of Eq.~\eqref{eq:m_field}, and $w_{0}=w_{1}=w_{2}=w_{3}=0$.

%% file: Sections/04_exp.tex
\section{Experiments}
\label{sec:experiments}
In this section, we first introduce datasets and evaluation metrics. Then, we conduct a correlation analysis to demonstrate that \method can measure human motion plausibility. Next, we evaluate our proposed motion prior on different tasks including denoising motion sequences, fitting to partial observations, jitter mitigation for SMPL-based human pose estimators and refining the results of motion in-betweening. Additionally, in our supplementary document, we conducted the extended experiment on \textbf{refining motion in-betweening results}, the experiment on \textbf{motion generation}, as well as \textbf{ablation studies} on motion segment length, different temporal optimization terms, only using our proposed prior and loss functions.
Please refer to the supplementary for more qualitative results and experiment details.

\boldparagraph{Datasets.}
Following~\cite{tiwari22posendf,zeng2022smoothnet,qin2022motion}, we evaluate our motion prior on five datasets including AMASS~\cite{AMASS:ICCV:2019}, HPS~\cite{HPS}, 3DPW~\cite{vonMarcard2018}, AIST++~\cite{li2021learn} and LAFAN1~\cite{harvey2020robust}. For 3DPW and AIST++, we utilize the data organized by SmoothNet~\cite{zeng2022smoothnet}, consisting of the results of various human pose estimators. For detailed datasets description, please refer to our supplementary.

\boldparagraph{Evaluation Metrics.}
In the evaluation, five standard metrics are used, including the mean per joint position error (MPJPE), the Procrustes-aligned mean per joint position error (PA-MPJPE), the mean per vertex position error (PVE), the acceleration error (Accel), and normalized power spectrum similarity (NPSS). For more detailed description, please refer to our supplementary material.
\ssecspace
\subsection{Correlation Analysis}
\label{subsec:interpretability}
\input{tables_images_tex/pearson}
In this section, we aim to validate whether the proposed \method is able to measure the human motion plausibility. To this end, we utilize the VIBE estimation on the 3DPW dataset and obtain 60,752 motion segments, for which we compute acceleration vectors' distances using the method from Sec.~\ref{subsec:loss}. We then calculate the acceleration error (\ie, the evaluation metric Accel) for each segment against 3DPW's ground truth. We measure the correlation between the manifold distances and acceleration error across joints using the Pearson correlation coefficient. The result is shown in Table~\ref{tab:pearson}. We can see that the proposed manifold distance for each joint has a strong correlation with the acceleration error. To be noted that, the distances are obtained by searching for the top-k nearest neighbors in AMASS (zero-level), while the acceleration error is calculated against the 3DPW ground-truths. This demonstrates our manifolds are sufficient to describe general motion patterns and the distances can also be used as a measure of motion smoothness and consistency with ground-truth motion. Please refer to our supplementary material for more intuitive visualization of the positive linear correlation.

\ssecspace
\subsection{Motion Denoising}
\label{subsec:motion_denoise}
In this section, we conduct the motion denoising experiment, which aims to enhance the quality of captured motion sequences through an optimization-based method, with the goal of aligning the recovered human body well with the observations and preserving the realism of human poses and motion. We compared our \method with the SOTA method Pose-NDF~\cite{tiwari22posendf} and two other pose or motion priors VPoser-t~\cite{SMPL-X:2019} and HuMoR~\cite{rempe2021humor}. All the results of Pose-NDF are obtained using their released code and model.

For a fair comparison, we follow the setup of Pose-NDF, disregarding the translation and global orientation of the root joint. Similar to the optimization objective in Pose-NDF, we replace the temporal term, \ie, Eq.~\eqref{eq:posendf-smooth}, in Pose-NDF with our motion prior term, \ie,~Eq.~\eqref{eq:fusion_term}:
\begin{equation}
  \varepsilon_{temp}^{posendf}=\left \| M(\boldsymbol{\beta}_{0},\boldsymbol{\theta} ^{t})- M(\boldsymbol{\beta}_{0},\boldsymbol{\theta} ^{t-1}) \right \| _{2}^{2},
\label{eq:posendf-smooth}
\end{equation}
where $M(\boldsymbol{\beta},\boldsymbol{\theta})$ represents SMPL mesh vertices for the given pose ($\boldsymbol{\theta}$) and shape ($\boldsymbol{\beta}$) parameters of SMPL model. Thus, we find the pose parameter $\boldsymbol{\theta}^{t}$ at frame t with:
\begin{equation}
\boldsymbol{\theta}^{t}=\underset{\boldsymbol{\theta}}{\text{arg min}}~\lambda_{v}\varepsilon_{v}+\lambda_{\boldsymbol{\theta}}\varepsilon_{\boldsymbol{\theta}}+\lambda_{f}\varepsilon_{fusion},
\label{eq:motion-denoise}
\end{equation}
where \ziqiang{$\lambda_{v},\lambda_{\boldsymbol{\theta}},\lambda_{f}$ are the optimization weights}, $\varepsilon_{v}$ makes sure that the optimized pose is close to the observation and the pose prior term $\varepsilon_{\boldsymbol{\theta}}$ keeps the pose plausible:
\begin{equation}
\varepsilon_{v}=\left \| \mathcal{J}(\boldsymbol{\beta}_0,\boldsymbol{\theta}^{t})-\mathcal{J}_{obs} \right \|_{2}^{2}~~~~~~\varepsilon_{\boldsymbol{\theta}}=f_{posendf}(\boldsymbol{\theta}), 
\label{eq:obs}
\end{equation}
where $\mathcal{J}_{obs}$ represents vertices or joints (mocap markers) and $f_{posendf}$ represents the pose prior learned by Pose-NDF. Finally, we use \method as a motion prior term \ie,~Eq.~\eqref{eq:fusion_term} in the optimization to preserve reasonable motion.

Following Pose-NDF, we create random noisy sequences by adding Gaussian noise to two mocap datasets (HPS and test split of AMASS) and name them "Noisy HPS" and "Noisy AMASS" respectively. The average noise introduced in "Noisy HPS" is 8.7 cm and "Noisy AMASS" is 9.0 cm. Similar to Pose-NDF, we create the data with a fixed shape and do not optimize the shape parameters $\boldsymbol{\beta}$. 

\input{tables_images_tex/denoise_partial}

\footnotetext{For the 240-frame experiment of HuMoR, all sequences crashed and could not be effectively denoised due to error accumulation, therefore, there is no data here.}
For VPoser-t, we use VPoser as the pose prior, and employ the temporal term in the latent space to smooth the motion like~\cite{tiwari22posendf,zhang2021learning}, and we optimize the latent code of poses in the VAE-based latent space. Thus, the pose prior and temporal term are given as:
\begin{equation}
\varepsilon_{\boldsymbol{\theta}}^{VPoser-t}=\left \| z^{t} \right \| _{2}~~~~~~\varepsilon_{temp}^{VPoser-t}=\left \| z^{t-1}-z^{t} \right \| _{2},
\label{eq:vposer}
\end{equation}
where $z^{t}$ is the latent code of the pose~$\boldsymbol{\theta}^{t}$ encoded by the VPoser. 
We start the optimization from the same initial poses for a fair comparison.

The experimental results are shown in Table~\ref{tab:motiondenoise}. 
We can see that our method consistently achieves the lowest errors across all settings. This superior performance can be attributed to its enhanced capability to model human dynamics better than existing pose or motion priors.
By modeling human motion as a neural distance field, we can explicitly quantify human dynamics as a distance value, which can serve as a metric to guide the optimization process. This modeling is performed in the continuous space, departing from the previous approaches which were often conducted in the biased Gaussian spaces of VAE-based representations.

\subsection{Fitting to Partial Data}
\label{subsec:parial_fitting}
In this section, we conduct the experiment of fitting to partial data where some joints are occluded, meaning that, there are no corresponding observations in Eq.~\eqref{eq:obs}. We use the test set of AMASS to perform this experiment under three different occlusion scenarios: arm, leg, and shoulder. We randomly select some frames from motion sequences and designate their corresponding body joints as occluded to create occluded poses and quantitatively
compare with the SOTA Pose-NDF and two other pose or motion priors VPoser-t and HuMoR. 
For VPoser-t and HuMoR, we use the same optimization objectives as described in Sec.~\ref{subsec:motion_denoise}.

Because \method can better preserve human motion dynamics, our method outperforms others in all cases as shown in Table~\ref{tab:partial}. In contrast, HuMoR encounters issues with error accumulation over time due to the modeling of transitions between only two consecutive frames, which has also been demonstrated  in~\cite{tiwari22posendf}.


\subsection{Mitigating Jitters for SMPL-based Pose Estimators}
\label{subsec:smooth}
\input{tables_images_tex/smpl_refine}

\footnotetext{Because the data on AIST++ organized by SmoothNet is partial, we only evaluate for video-based estimators.}
Human pose and shape estimation has broad applications such as avatar animation and human-computer interaction. Existing video-based pose estimators or image-based pose estimators when applied to videos often suffer from severe jitters, caused by rarely seen or occluded actions. As a motion prior, \method can be utilized to optimize the results of pose estimators to mitigate jitter issues and obtain more realistic motion. Here, we compare with the current SOTA method SmoothNet~\cite{zeng2022smoothnet} on the SMPL-based pose estimators.
\footnote{For SmoothNet, we use the model trained on 3D keypoints because they demonstrate that such models perform better than models trained on SMPL parameters. To ensure a fair comparison of generalization, we use the SPIN-3DPW model presented in their paper. We use their released model and data for comparison.}

The results are listed in Table \ref{tab:smooth} and Table \ref{tab:aist}.
The experimental results demonstrate that our approach achieves more accurate pose estimation while reducing acceleration error. 
In Table \ref{tab:smooth}, we also show the strong generalization performance of our motion prior. Notably, we do not utilize ground-truth annotations from any human pose estimation datasets except for the reasonable motion of AMASS dataset. We rely on the learned motion prior to optimize the results of pose estimators with Eq.~\eqref{eq:fusion_term}. 
We also use a simple while effective moving average~\cite{hunter1986exponentially} strategy to smooth the global orientation, which differs from SmoothNet~\cite{zeng2022smoothnet}.
For the TCMR~\cite{choi2021beyond}, since it has used some smoothing strategies in its model, we use the Eq.~\eqref{eq:mo_term} to optimize the results.
Additionally, we segment the human body mesh to compute PVE for leg, foot and toe-base (in mm) in Table~\ref{tab:footskate}. It can be seen that, for video-based estimators, our method avoids excessive motion smoothing, unlike SmoothNet, which may cause unnatural leg and foot movements,~\ie, footskate, as shown in Fig.~\ref{fig:vsSmoothNet}. 

\subsection{Motion In-betweening Refinement}
\label{subsec:inbetween}
\input{tables_images_tex/vssmoothnet_image}
\begin{figure*}[!t]
    \addtolength{\belowcaptionskip}{-0.2cm}
    \centering
    \includegraphics[width=1\linewidth, trim={0 0 0 10mm}, clip]{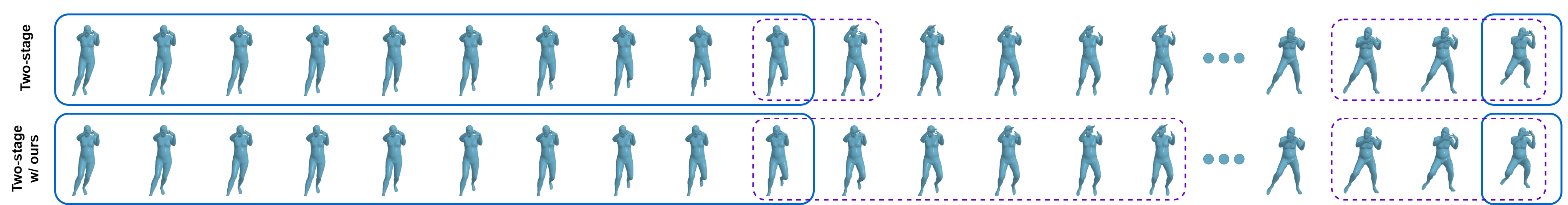}
    \vspace{-1.7em}
    \caption{
    \textbf{Qualitative Comparison with Two-stage.} The poses in blue boxes are the initial poses and the target pose. And the intermediate poses are the generated transitions. In the first row, Two-stage produces unnatural transitions shown in the purple boxes, and the hands and legs undergo sudden changes, which is obviously inconsistent with the motion trends before and after. While after our optimization, more natural transitions can be achieved.
    }
    \label{fig:inbetween}
\end{figure*}
Motion in-betweening aims to generate natural intermediate frames between initial and target poses.
Although extensive progress has been made, existing methods may still generate some unnatural transitions because human motion is inherently complex and stochastic. In this section, we utilize our motion prior (without the traditional temporal term), \ie, Eq.~\eqref{eq:mo_term}, to further optimize the results of current SOTA method Two-stage~\cite{qin2022motion} in order to obtain more natural transitions. The results are shown in Table~\ref{tab:inbetween} and Fig.~\ref{fig:inbetween}. We can see that our \method demonstrates good generalization and improves the results of existing SOTA learning-based method
by reducing acceleration error and producing more lifelike motion.

%% file: tables_images_tex/pearson.tex
\begin{table}
  \begin{minipage}[c]{0.5\textwidth}
    \resizebox{\linewidth}{!}{
    \centering
    \scriptsize
    \addtolength{\belowcaptionskip}{-0.2cm}
    \setlength{\extrarowheight}{1pt} 
    \begin{tabular}{>{\centering}p{1.2cm}>{\centering}p{0.8cm}|>{\centering}p{1.2cm}>{\centering}p{0.8cm}|cc}
    \hline
        \textbf{Joint} & \textbf{Pearson}& \textbf{Joint} & \textbf{Pearson} & \textbf{Joint} & \textbf{Pearson} \\ \hline
        leftKnee & 0.9869 & neck & 0.9631 & rightElbow & 0.9913 \\
        rightKnee & 0.9863 & leftCollar & 0.9455 & leftWrist & 0.9946 \\
        leftAnkle & 0.9890 & rightCollar & 0.9638 & rightWrist & 0.9946 \\
        rightAnkle & 0.9886 & leftShoulder & 0.9700 & leftHand & 0.9908 \\
        leftFoot & 0.9829 & rightShoulder & 0.9759 & rightHand & 0.9910 \\
        rightFoot & 0.9810 & leftElbow & 0.9903 & ~ & ~ \\ \hline
    \end{tabular}
    }
  \end{minipage}\hfill
  \begin{minipage}[c]{0.48\textwidth}
    \caption{
    \textbf{Pearson Correlation Coefficient.} All Pearson coefficients are very close to 1, indicating strong linear correlations between manifold distances and acceleration error. 
    } \label{tab:pearson}
  \end{minipage}
  \vspace{-0.4cm}
\end{table}

%% file: tables_images_tex/denoise_partial.tex
\begin{table}
\addtolength{\belowcaptionskip}{-0.3cm}
    \begin{minipage}[t]{0.48\linewidth}
    \vspace{-2.4em}
    \begin{center}
    \resizebox{\linewidth}{!}{
        \tiny
        \centering
        \setlength{\tabcolsep}{8pt}
        \setlength{\extrarowheight}{3pt} 
        \begin{tabular}{c>{\centering}p{0.3cm}>{\centering}p{0.45cm}>{\centering}p{0.35cm}>{\centering}p{0.45cm}>{\centering}p{0.45cm}c} 
        \hline
        \textbf{Data}       & \multicolumn{3}{c}{\textbf{Noisy HPS}} & \multicolumn{3}{c}{\textbf{Noisy AMASS}} \\ \cline{2-7} 
        \textbf{\# frames}  & 60     & 120    & 240     & 60     & 120    & 240     \\ \hline{}
        VPoser-t~\cite{SMPL-X:2019} & 3.05 & 4.43 & 7.11 & 5.83 & 6.55 & 7.86 \\
        HuMoR~\cite{rempe2021humor} & 6.08 & 12.67 & - & 10.28 & 12.63 & - \\
        Pose-NDF~\cite{tiwari22posendf} & 1.17 & 1.30 & 1.16 & 5.03 & 5.39 & 5.49 \\
            \textbf{Ours} & \textbf{0.90} & \textbf{0.91} & \textbf{0.88} & \textbf{1.45} & \textbf{1.47} & \textbf{1.60} \\ \hline
        \end{tabular}
    }
    \end{center}
    \vspace{-0.5em}
    \caption{\textbf{Motion Denoising.} We compare PVE in cm. \protect\footnotemark} 
    \label{tab:motiondenoise}
    \end{minipage}
    \hfill
    \begin{minipage}[t]{0.48\linewidth}
    \begin{center}
    \resizebox{\linewidth}{!}{
    \setlength{\tabcolsep}{8pt}
    \tiny
    \centering
    \setlength{\extrarowheight}{1.5pt} 
    \begin{tabular}{ccccccc}
    \hline
    \multirow{2}{*}{\textbf{Data}} & \multicolumn{2}{c}{\multirow{2}{*}{\textbf{Occ. Leg}}} & \multicolumn{2}{c}{\textbf{Occ. Arm}}            & \multicolumn{2}{c}{\textbf{Occ. Shoulder}}       \\
                          & \multicolumn{2}{c}{}                          & \multicolumn{2}{c}{\textbf{+Hand}}               & \multicolumn{2}{c}{\textbf{+Upper Arm}}          \\ \cline{2-7} 
    \textbf{\# frames}               & 60                    & 120                   & 60            & \multicolumn{1}{c}{120} & 60            & \multicolumn{1}{c}{120} \\ \hline
    VPoser-t~\cite{SMPL-X:2019}      & 8.69                  & 10.77                 & 8.79          & 10.70                   & 8.74          & 10.20                   \\
    HuMoR~\cite{rempe2021humor}      & 9.52                  & 12.70                 & 9.39          & 13.82                   & 9.02          & 12.14                   \\
    Pose-NDF~\cite{tiwari22posendf}  & 8.50                  & 9.40                  & 8.66          & 9.43                    & 8.73          & 9.47                    \\
    \textbf{Ours}                    & \textbf{4.83}         & \textbf{5.07}         & \textbf{4.83} & \textbf{5.01}           & \textbf{4.93} & \textbf{5.04}           \\ \hline
    \end{tabular}
    }
    \end{center}
    \vspace{-0.5em}
    \caption{\textbf{Fitting to Partial Data.} We compare PVE (in cm) on test set of AMASS. 
    }
    \label{tab:partial}
    \end{minipage}
\end{table}

%% file: tables_images_tex/smpl_refine.tex
\begin{table}
    \begin{minipage}[t]{0.46\linewidth}
    \begin{center}
    \resizebox{\linewidth}{!}{
        \tiny
        \centering
        \setlength{\tabcolsep}{8pt}
        \setlength{\extrarowheight}{1.5pt} 
        \begin{tabular}{c>{\centering}p{0.75cm}>{\centering}p{1.25cm}>{\centering}p{0.6cm}c}
        \hline
        \multicolumn{1}{c}{\multirow{2}{*}{\textbf{Method}}} & \multicolumn{4}{c}{\textbf{3DPW}} \\ \cline{2-5} 
        \multicolumn{1}{c}{} & \textbf{MPJPE $\downarrow$} & \textbf{PA-MPJPE $\downarrow$} & \textbf{PVE $\downarrow$} & \textbf{Accel $\downarrow$} \\ \hline
        SPIN~\cite{kolotouros2019spin} & 99.29 & 61.71 & 113.32 & 34.95 \\
        SPIN w/ S~\cite{zeng2022smoothnet} & 97.81 & 61.19 & 111.5 & \textbf{7.4} \\
        SPIN w/ ours & \textbf{97.24} & \textbf{60.80} & \textbf{111.37} & 8.43 \\ \hline
        EFT~\cite{joo2020eft} & 91.6 & 55.33 & 110.17 & 33.38 \\
        EFT w/ S~\cite{zeng2022smoothnet} & 89.57 & 54.40 & \textbf{107.66} & \textbf{7.89} \\
        EFT w/ ours & \textbf{89.35} & \textbf{53.83} & 107.82 & 8.94 \\ \hline
        PARE~\cite{Kocabas_PARE_2021} & 79.93 & 48.74 & 94.07 & 26.45 \\
        PARE w/ S~\cite{zeng2022smoothnet} & 78.68 & 48.47 & \textbf{92.5} & \textbf{6.31} \\
        PARE w/ ours & \textbf{78.55} & \textbf{47.84} & 92.65 & 7.63 \\ \hline
        VIBE*~\cite{kocabas2019vibe} & 84.28 & 54.93 & 99.10 & 23.59 \\
        VIBE* w/ S~\cite{zeng2022smoothnet} & 83.46 & 54.83 & 98.04 & \textbf{7.42} \\
        VIBE* w/ ours & \textbf{83.07} & \textbf{54.28} & \textbf{97.8} & 8.01 \\ \hline
        TCMR*~\cite{choi2021beyond} & 88.47 & 55.70 & 103.22 & 7.13 \\
        TCMR* w/ S~\cite{zeng2022smoothnet} & 88.69 & 56.61 & 103.40 & \textbf{6.48} \\
        TCMR* w/ ours & \textbf{88.28} & \textbf{55.69} & \textbf{103.02} & 6.72 \\ \hline
        \end{tabular}
    }
    \end{center}
    \vspace{-0.5em}
    \caption{\textbf{Mitigating Jitters on 3DPW.} 
    \mbox{"w/ S"} indicates using SmoothNet. "*" denotes spatio-temporal backbones.}
    \label{tab:smooth}
    \end{minipage}
    \hfill
    \begin{minipage}[t]{0.52\linewidth}
    \begin{center}
    \resizebox{\linewidth}{!}{
    \setlength{\tabcolsep}{8pt}
    \tiny
    \centering
    \setlength{\extrarowheight}{1.5pt} 
    \begin{tabular}{c>{\centering}p{0.75cm}>{\centering}p{1.25cm}>{\centering}p{0.6cm}c}
    \hline
    \multicolumn{1}{c}{\multirow{2}{*}{\textbf{Method}}} & \multicolumn{4}{c}{\textbf{AIST++}} \\ \cline{2-5} 
    \multicolumn{1}{c}{} & \textbf{MPJPE $\downarrow$} & \textbf{PA-MPJPE $\downarrow$} & \textbf{PVE $\downarrow$} & \textbf{Accel $\downarrow$} \\ \hline
    VIBE*~\cite{kocabas2019vibe} & 107.41 & 72.83 & 127.56 & 31.65 \\
    VIBE* w/ S~\cite{zeng2022smoothnet} & 105.21 & \textbf{70.74} & 124.78 & \textbf{6.34} \\
    VIBE* w/ ours & \textbf{104.85} & 71.60 & \textbf{124.60} & 7.92 \\ \hline
    TCMR*~\cite{choi2021beyond} & 106.95 & 71.58 & 124.73 & 6.47 \\
    TCMR* w/ S~\cite{zeng2022smoothnet} & 107.19 & \textbf{71.43} & 124.76 & \textbf{4.70} \\
    TCMR* w/ ours & \textbf{106.51} & 71.56 & \textbf{124.20} & 5.29 \\ \hline
    \end{tabular}
    }
    \end{center}
    \vspace{-0.5em}
    \caption{\textbf{Mitigating Jitters on AIST++.}\protect\footnotemark}
    \label{tab:aist}
    
    \vspace{-0.25cm}
    
    \begin{center}
    \resizebox{\linewidth}{!}{
    \setlength{\tabcolsep}{8pt}
    \tiny
    \centering
    \setlength{\extrarowheight}{1.5pt} 
    \begin{tabular}{c>{\centering}p{0.4cm}>{\centering}p{0.4cm}>{\centering}p{0.5cm}>{\centering}p{0.5cm}>{\centering}p{0.4cm}c}
    \hline
    \multirow{3}{*}{\textbf{Method}} & \multicolumn{6}{c}{\textbf{3DPW}}                                                                 \\ \cline{2-7} 
                                     & \multicolumn{3}{c}{\textbf{Left-}}              & \multicolumn{3}{c}{\textbf{Right-}}             \\
                                     & \textbf{Leg} & \textbf{Foot} & \textbf{ToeBase} & \textbf{Leg} & \textbf{Foot} & \textbf{ToeBase} \\ \hline
    VIBE*~\cite{kocabas2019vibe} & 99.73 & 137.11 & 144.40 & 101.13 & 139.86 & 149.85 \\
    VIBE* w/ S~\cite{zeng2022smoothnet} & 99.76 & 137.43 & 144.50 & 101.20 & 140.40 & 150.00 \\
    VIBE* w/ ours & \textbf{98.66} & \textbf{136.21} & \textbf{143.65} & \textbf{99.86} & \textbf{138.70} & \textbf{148.93} \\ \hline
    TCMR*~\cite{choi2021beyond} & 99.72 & 140.21 & 148.60 & 101.94 & 142.29 & \textbf{152.56} \\
    TCMR* w/ S~\cite{zeng2022smoothnet} & 100.19 & 141.18 & 149.48 & 103.05 & 144.25 & 154.31 \\
    TCMR* w/ ours & \textbf{99.54} & \textbf{140.02} & \textbf{148.42} & \textbf{101.84} & \textbf{142.28} & 152.57 \\ \hline
    \end{tabular}
    }
    \end{center}
    \vspace{-0.5em}
    \caption{\textbf{Mitigating Jitters of Legs and Feet.}
    }
    \label{tab:footskate}
    
    \end{minipage}
\vspace{-0.2cm}
\end{table}

%% file: tables_images_tex/vssmoothnet_image.tex

\begin{figure}
  \begin{minipage}[c]{0.46\textwidth}
    \centering
    \includegraphics[width=\textwidth]{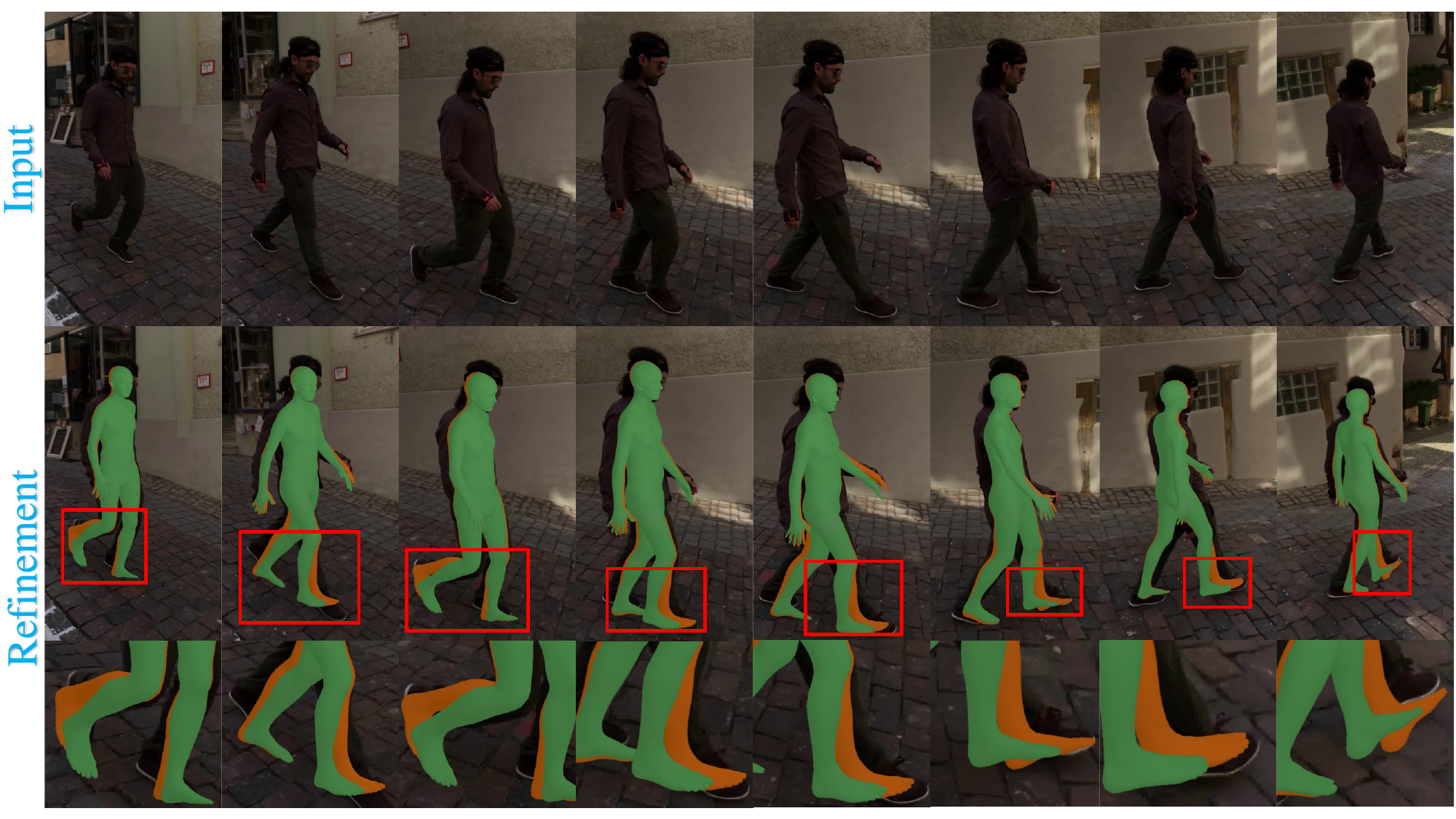}
  \end{minipage}\hfill
  \begin{minipage}[c]{0.52\textwidth}
    \caption{
    \textbf{Qualitative Comparison.} 
    We refine the estimation results of VIBE on 3DPW using SmoothNet (Green) and our method (Orange).
    When observing the video, it is apparent that SmoothNet will overly smooth the motion, making a walking person appear to skate.     
    In contrast our approach can well preserve human motion dynamics while mitigating the jitters issue.
    } \label{fig:vsSmoothNet}
  \end{minipage}
  \vspace{-0.2cm}
\end{figure}

%% file: Sections/05_conclusion.tex
\section{Conclusion}
\label{sec:Conclusion}
\input{tables_images_tex/inbetween}
This paper presents a novel human motion prior \method that models plausible human motions in continuous high-dimensional motion space with decoupled joint acceleration manifolds.
Extensive experiments demonstrate that \method has good generalization ability and outperforms existing SOTAs on multiple motion-related tasks.
Although the relationship between joints is implicitly established through the SMPL tree structure, 
such relationship is relatively weak. 
Therefore, as future work, we will explore how to establish explicit relationships between joints under the 
representation of neural distance field.


%% file: tables_images_tex/inbetween.tex
\begin{table}[!t]
  \begin{minipage}[c]{0.46\textwidth}
    \resizebox{\linewidth}{!}{
    \centering
    \scriptsize
    \setlength{\extrarowheight}{1.2pt} 
    \begin{tabular}{c>{\centering}p{0.4cm}>{\centering}p{0.4cm}>{\centering}p{0.4cm}ccc}
    \hline
        \textbf{Method} & \multicolumn{3}{c}{\textbf{NPSS $\downarrow$}} & \multicolumn{3}{c}{\textbf{Accel $\downarrow$}} \\ \cline{2-7}
        \textbf{frames} & \textbf{15} & \textbf{30} & \textbf{45} & \textbf{15} & \textbf{30} & \textbf{45} \\ \hline
        Two-stage~\cite{qin2022motion} & 0.06 & 0.28 & 0.68 & 11.97 & 11.10 & 10.47 \\
        Two-stage w/ ours & 0.06 & 0.28 & 0.68 & \textbf{7.71} & \textbf{8.03} & \textbf{8.08} \\ \hline    
    \end{tabular}
    }
  \end{minipage}\hfill
  \begin{minipage}[c]{0.52\textwidth}
    \caption{
    \textbf{Refining Motion In-betweening on LAFAN1.} "frames" refers to the number of frames of the generated transitions. 
    } \label{tab:inbetween}
  \end{minipage}
  \vspace{-0.2cm}
\end{table}

%% file: Sections/Outline.tex
\secspace
\vspace{-0.2cm}
\section*{Outline}
Here we provide details, extended experiments and ablation studies omitted from the main paper for brevity. App.~\ref{sec:details} provides implementation details, App.~\ref{sec:Evaluation_Details} gives the experimental evaluation details, App.~\ref{sec:Extended_Experiments} presents more experiments of our motion prior, App.~\ref{sec:ablation} contains our ablation studies and App.~\ref{sec:discuss} provides some extended discussions.
We encourage the reader to view the supplementary video for more qualitative results.

%% file: Sections/A_implementation_details.tex
\section{Implementation Details}
\label{sec:details}
\subsection{Weighted Design}
Here, we introduce the weighted design of Eq.~(\textcolor{red}{3}) in the main paper, where $w_{i}$ is determined by the summation of bone lengths from joint $i$ to the root joint along the kinematic structure of the SMPL body model. For joint $i$, the summation of bone lengths $l_{i}$ is,
\vspace{-0.2cm}
\begin{equation}
  l_{i} = \sum b,
\label{eq:path}
\vspace{-0.2cm}
\end{equation}
where b is the bone length. Thus, through experimental exploration, $w_{i}$ is defined as:
\vspace{-0.2cm}
\begin{equation}
  w_{i} = \frac{4l_{i}^{2}}{4l_{i}^{2}+1}.
\label{eq:wi}
\vspace{-0.2cm}
\end{equation}
This design ensures that joints with intenser movements contribute proportionally more to the unsigned distance field of motion segment $\mathbf{m}$.

\subsection{Data Preparation}
\boldparagraph{Training Data.}
The training data is divided into two categories: plausible motion data and noisy motion data. We use the train split of AMASS dataset~\cite{AMASS:ICCV:2019} as the plausible motion data,~\ie, the zero level of plausible acceleration vectors manifolds. We downsample AMASS to 25Hz or 24Hz because it records human motion at 100Hz or 120Hz. This will ensure that the temporal gap of consecutive frames between the two frequency motion data is closest and it can be easily generalized to higher frequencies \eg, 30Hz. Then, we randomly sample motion segments of fixed lengths to model the manifolds. 

For the noisy motion data, which lies outside the manifold, we utilize artificially noised motion data and the results from a representative SMPL-based human pose estimator VIBE~\cite{kocabas2019vibe}. We apply the noise from a uniform distribution, rather than Gaussian noise, to create artificially noised motion data of AMASS training set. Because it will produce a more diverse and wider distribution of noisy motion. Specifically, for a motion segment of length $y$, we randomly select $x~(x\le y)$ frames for adding noise, where $x$ is also randomly generated. Furthermore, after employing manually noised motion data for training, we performed fine-tuning using the results from the human pose estimator VIBE on videos of MPI-INF-3DHP dataset~\cite{mono-3dhp2017}. Due to self-occlusion and partial observations, the estimates output by existing estimators encompass a substantial amount of noisy motion that is hard to be replicated through artificial noise. Additionally, such noisy motion is closer to the manifold, which will help learn a more refined manifold surface. Notably, we do not use any ground truth annotations
from the MPI-INF-3DHP dataset.

We employ KNN algorithm~\cite{cover1967nearest} to compute the ground truth distance values of acceleration vectors outside the manifold. We implement KNN using FAISS~\cite{johnson2019billion}. Specifically, for an acceleration vector, we calculate the top-k nearest distances to the zero level and then compute the average distance as the ground truth distance. In our setup, we use $k=5$.

\boldparagraph{Evaluation Data.}
For the motion denoising experiments in Sec.~\textcolor{red}{4.2}, we utilize two real world
mocap data HPS~\cite{HPS} and the test split of AMASS~\cite{AMASS:ICCV:2019}. HPS records human motion at 30Hz, thus we do not perform downsampling and directly conduct the evaluation on the motion of 30Hz. However, for the AMASS dataset, which records human motion at 100Hz or 120Hz, we downsample it to 25Hz or 24Hz for the evaluation. For HPS dataset, we randomly sampled 150 motion sequences for each setup. And in the experiments of AMASS dataset, we randomly selected 100 motion sequences for 60 frames or 120 frames. However, for the 240 frames of AMASS, due to downsampling requirements, we could only randomly sample 71 motion sequences for evaluation. Then, following Pose-NDF~\cite{tiwari22posendf}, we introduce random noise to each frame to create noisy observations.

In the fitting to partial experiments of Sec.~\textcolor{red}{4.3}, we use the test split of AMASS for evaluation, which is also downsampled to 25Hz or 24Hz. We also randomly selected 100 motion sequences for 60 frames or 120 frames. 
To simulate occlusion, we randomly select one-third of the frames within a motion sequence and set the rotations of corresponding occluded joints to zero. Besides, during optimization, when calculating the observation alignment term \ie, Eq.~(\textcolor{red}{12}) in main paper, the occluded joints of occluded frames are excluded.
\subsection{Optimization Details}
Since our motion prior is built upon motion segments, for an entire motion sequence, we initially split it into distinct motion segments by employing a sliding window with the window size equal to the length of our prior and the stride of $1$. This will avoid boundary effects and make any motion segment comply with human motion dynamics.
Subsequently, we calculate the distance of each motion segment to the plausible motion manifold, and then utilize the average distance of these motion segments to guide the optimization process.
For the experiments of motion denoising and fitting to partial observations, the optimization variable in Adam~\cite{kingma2014adam} is the entire motion sequence.
Specially, for post-optimization of human pose estimators and motion in-betweening, as we only use our motion prior without any other optimization objectives, we optimize each motion segment individually, recording multiple results of each frame to obtain the final optimized poses with a weighted average strategy similar to SmoothNet. It will increase the receptive field of each frame during optimization.

%% file: Sections/B_Experimental_Evaluation_Details.tex
\section{Experimental Evaluation Details}
\label{sec:Evaluation_Details}
\subsection{Datasets}
We evaluate our motion prior on five datasets including AMASS~\cite{AMASS:ICCV:2019}, HPS~\cite{HPS}, 3DPW~\cite{vonMarcard2018}, AIST++~\cite{li2021learn} and LAFAN1~\cite{harvey2020robust}.

AMASS is a large motion capture database containing diverse motion and body shapes on the SMPL body model. We sub-sample the dataset to 25Hz or 24Hz and use the recommended training split to train the unsigned distance fields. For the evaluation data, we also perform the same downsampling on the test split of AMASS.

HPS is a method to recover the full 3D pose of a human registered with a 3D scan of the surrounding environment using wearable sensors. And with this method, HPS recorded several large 3D scenes (300-1000 sq.m) consisting of 7 subjects and more than 3 hours of diverse motion.

3DPW is a challenging in-the-wild dataset consisting of 60 videos, which are captured by a phone at 30 FPS. Moreover, IMU sensors are utilized to obtain the near ground-truth SMPL parameters, \ie, pose and shape.

AIST++ is a challenging dataset that comes from the AIST Dance Video DB~\cite{aist-dance-db}. 
It contains 1,408 sequences of 3D human dance motion, represented as joint rotations along with root trajectories.

LAFAN1 is a high-quality public motion capture dataset. It contains 15 actions performed by 5 actors such as walking, dancing, fighting, jumping, with 496,672 frames captured in a production-grade motion capture system at 30Hz. We adopt the same test set in~\cite{qin2022motion}, which contains 2,232 clips sampled with a window of 65, offset by 40 frames on Subject 5. Although this dataset is not based on SMPL, its human skeleton definition is completely consistent with SMPL and joint rotations are provided, so the poses of this dataset can be converted into SMPL poses. In addition, since the rest-pose of this dataset is not T-Pose, the relative rotations of the joints in the dataset cannot be directly converted to those of SMPL, so we first converted the dataset so that all joint rotations are all relative to T-Pose.

\subsection{Evaluation Metrics}
For the evaluation, five standard metrics are used, including MPJPE, PA-MPJPE, PVE, and Accel.

MPJPE (Mean Per Joint Position Error) is calculated as the mean of the Euclidean distance between the ground-truth and the estimated 3D joint positions after aligning the pelvis joint on the ground truth location. MPJPE comprehensively evaluates the predicted poses and shapes, including the global orientations.

PA-MPJPE (Procrustes-Aligned Mean Per Joint Position Error) performs Procrustes alignment before computing MPJPE, which mainly measures the articulated poses, eliminating the differences in scale and global orientation.

PVE (Mean Per Vertex Position Error) is calculated as the mean of the Euclidean distance between the ground truth and the estimated 3D human mesh vertices (output by the SMPL model).

Accel (Mean Per Joint Acceleration Error) is measured as the mean difference between the ground-truth and the estimated 3D acceleration for every joint. It is used to express the smoothness and temporal coherence of 3D human motion as well as the similarity to ground-truth motion.

NPSS (The Normalized Power Spectrum Similarity) proposed by~\cite{gopalakrishnan2019neural},
evaluates angular differences between predicted motion and ground truth on the frequency domain. NPSS measures similarity of motion patterns, which reportedly correlates better with human perception of quality.

\ssecspace
\subsection{PVE of Legs and Feet}
\input{tables_images_tex/supp_smpl_corr}

In Table 5 of the main paper, we employ the PVE (Per Vertex Error) of legs and feet to numerically demonstrate that our method avoids resulting in footskate when smoothing the motion, compared with SmoothNet~\cite{zeng2022smoothnet}. As shown in Figure~\ref{fig:smplseg}, we segment the human body mesh into different parts through the indices of mesh vertices provided by ~\cite{SMPL:2015} and then compute the PVE for the vertices belonging to the legs, feet and toe-bases (in mm).
\ssecspace
\subsection{Optimization Space of Rotation}
For fair comparison, in Sec.~\textcolor{red}{4.2} and Sec.~\textcolor{red}{4.3}, we optimize the human poses in the axis-angle space same with Pose-NDF~\cite{tiwari22posendf}, and in Sec.~\textcolor{red}{4.4}, we adopt the space of 6D rotation representation~\cite{zhou2019continuity} following SmoothNet~\cite{zeng2022smoothnet}.
Moreover, we have observed that optimizing human poses in the 6D space is more stable and leads to better convergence in some cases compared to the axis-angle space. Therefore, in Sec.~\textcolor{red}{4.5} and Sec.~\ref{subsec:flexible}, we optimize the human poses in the 6D space.

%% file: tables_images_tex/supp_smpl_corr.tex
\begin{figure}
\addtolength{\belowcaptionskip}{-1em}
    \begin{minipage}[t]{0.48\linewidth}
    \begin{center}
    \resizebox{0.65\linewidth}{!}{
    \includegraphics[width=\linewidth, trim={0 0 0 0}, clip]{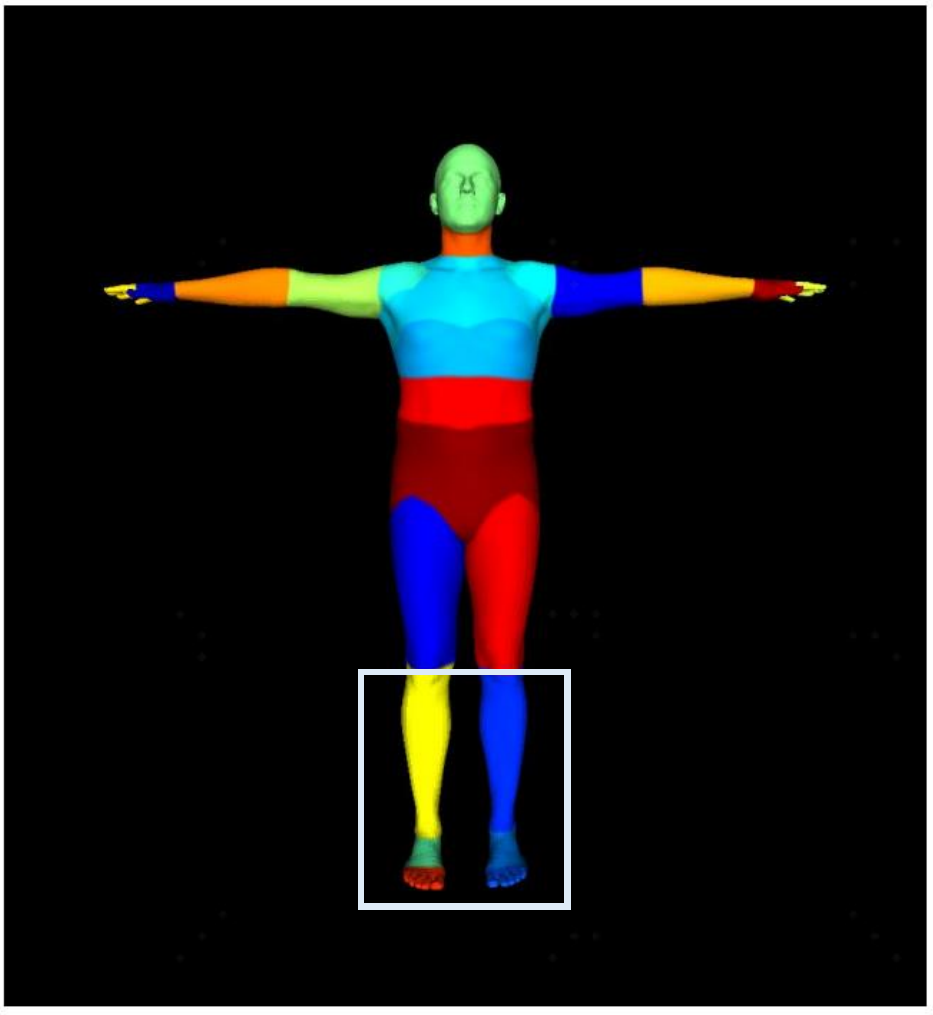}
    }
    \end{center}
    \vspace{-1.5em}
    \caption{\textbf{SMPL Body Segmentation.} The white box contains the segmentation of legs, feet and toe-bases.} 
    \label{fig:smplseg}
    \end{minipage}
    \hfill
    \begin{minipage}[t]{0.48\linewidth}
    \vspace{-12em} %
    \begin{center}
    \resizebox{\linewidth}{!}{
    \includegraphics[width=0.9\linewidth, trim={0 5mm 0 0}, clip]{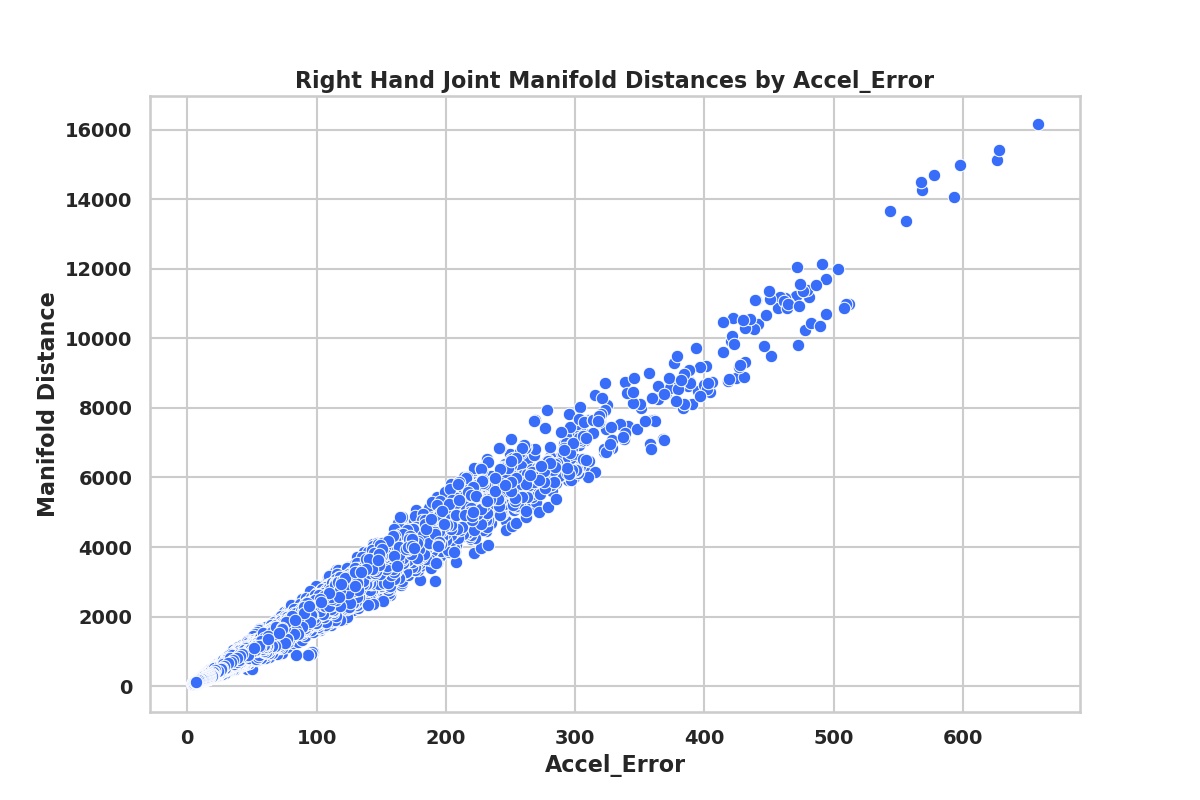}
    }
    \end{center}
    \vspace{-0.8em}
    \caption{
    \textbf{SMPL Body Segmentation.} The white box contains the segmentation of legs, feet and toe-bases.
    }
    \label{fig:correlation1}
    \end{minipage}
\end{figure}

%% file: Sections/C_Extended_Experiments.tex
\secspace
\section{Extended Experiments}
\label{sec:Extended_Experiments}
For dynamic motion and better qualitative comparison, we recommend viewing our supplementary video.
\ssecspace
\subsection{Correlation Analysis}
\begin{figure*}[!t]
    \addtolength{\belowcaptionskip}{-1.0em}
    \centering
    \includegraphics[width=1\linewidth, trim={0 5mm 0 0}, clip]{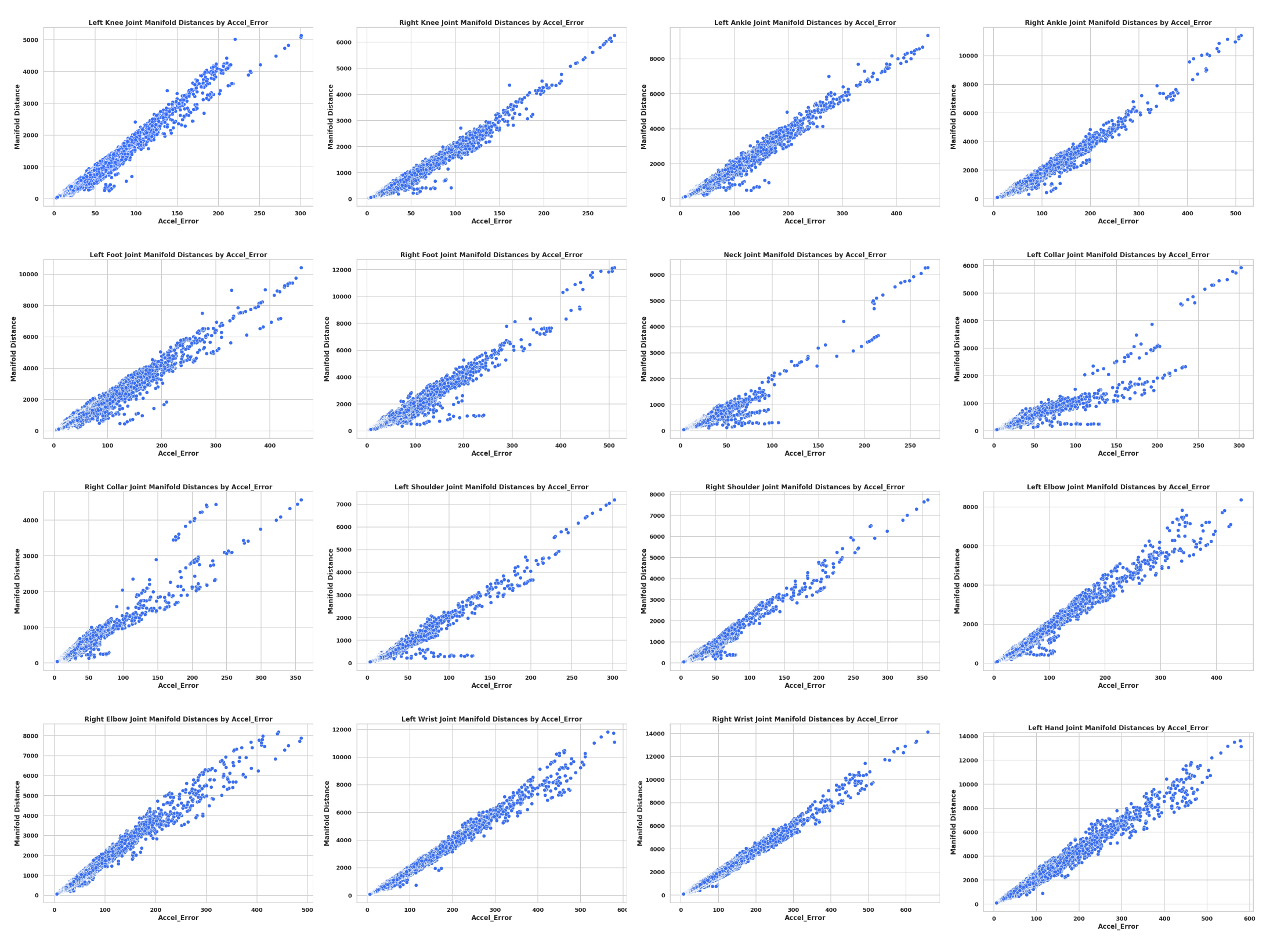}
    \vspace{-1.6em}
    \caption{
    \textbf{Scatter Plots of Other Joints.} Each blue point represents a motion segment.
    }
    \label{fig:correlation2}
\end{figure*}

In this section, we will present the intuitive visualization of the positive linear correlations between the manifold distances and acceleration error across joints. 
The linear correlation of the right hand joint are visualized in Figure~\ref{fig:correlation1}. Moreover, 
Figure~\ref{fig:correlation2} shows the linear correlations of the other joints. The two joints on the spine and the head joint are missing here because there are no corresponding joints in the GT skeleton of 3DPW, so acceleration error cannot be obtained.
\begin{figure}[!t]
    \addtolength{\belowcaptionskip}{-1.0em}
    \centering
    \includegraphics[width=1\linewidth, trim={0 0 0 0}, clip]{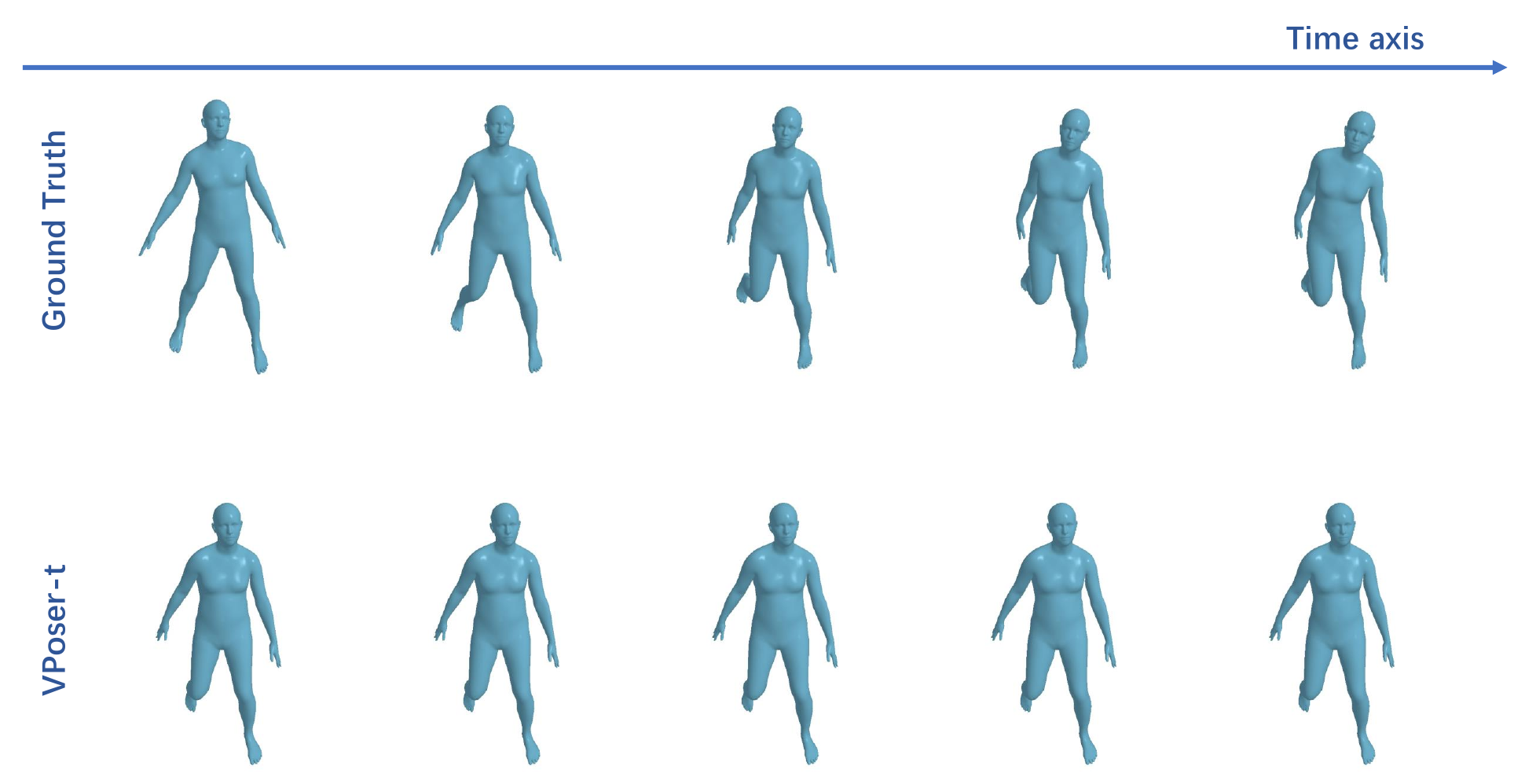}
    \vspace{-1.6em}
    \caption{
    \textbf{VPoser-t Denoising Results.} 
    }
    \label{fig:vposer}
\end{figure}

\subsection{Motion Fitting from 3D Observations}
\input{tables_images_tex/supp_motion_fitting}
\begin{figure*}[!t]
    \addtolength{\belowcaptionskip}{-1.5em}
    \centering
    \includegraphics[width=0.79\linewidth, trim={0 5mm 0 0}, clip]{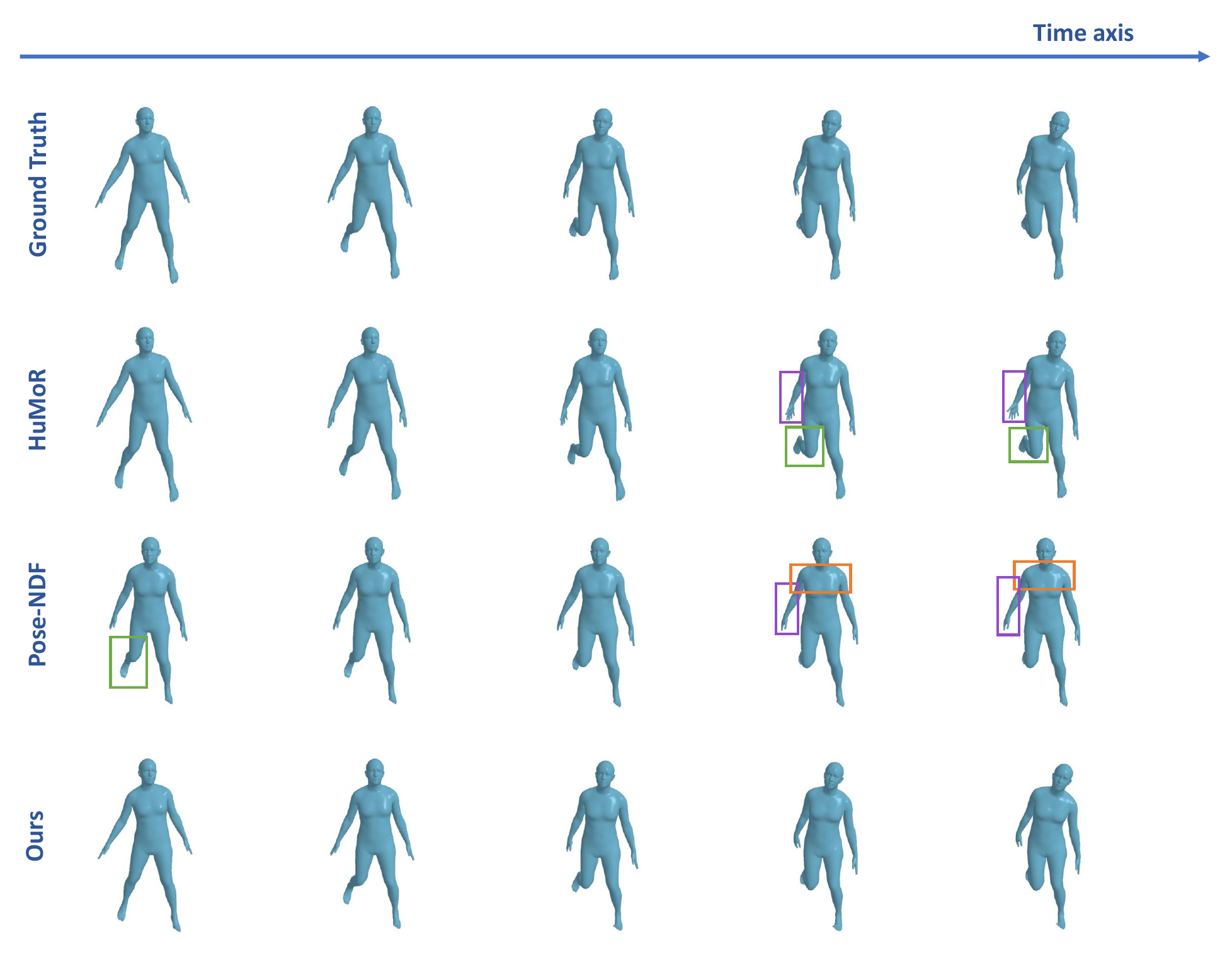}
    \vspace{-0.6em}
    \caption{
    \textbf{Denoising Comparison.} 
Body parts that are significantly different from the ground truth are marked in colored boxes. The results of VPoser-t are same with Figure~\ref{fig:vposer}. For uniform motion of Pose-NDF, the legs begin to retract in the first frame, whereas at that time, the human should stand on the ground. Besides, the right arm and shoulders in the last two frames are obviously different from the ground truth. Since this is the beginning of the motion, there is no accumulation of errors for HuMoR. And our results are the closest to the ground truth.
    }
    \label{fig:fusion}
\end{figure*}

VPoser-t~\cite{SMPL-X:2019} embeds human poses into a biased Gaussian space of VAE-based representations and optimizes poses within the latent space, resulting in average poses. When these average poses are assembled into motion, the resulting sequences appear stiff and mechanical as shown in Figure~\ref{fig:vposer}.

HuMoR~\cite{rempe2021humor} could also recover realistic motion in some cases, but due to modeling of
transitions between only two consecutive frames, there might be an accumulation of errors leading to extreme unrealistic poses (as shown in Figure~\ref{fig:humor}) in the final few frames of the motion, which has
also been demonstrated in~\cite{tiwari22posendf}.

Pose-NDF~\cite{tiwari22posendf} employs  a traditional temporal regularization term to smooth motion, but this tends to cause the uniform motion. Because the optimization direction of such temporal terms aims to minimize the frame-to-frame differences, effectively freezing the motion. Hence, the motion generated by Pose-NDF exhibits minimal variation in velocity, which will result in a lack of dynamism, particularly in actions that involve distinct changes in speed, such as pushing movements. Furthermore, the range of motion will also be restricted, as depicted in Figure~\ref{fig:range}.

In Figure~\ref{fig:fusion}, we present the initial five frames of the side hopping motion, and the results of our method are closest to ground truth since we can well preserve the human motion dynamics. We suggest watching our supplementary video for more qualitative results.
\subsection{Mitigating Jitters for SMPL-based Pose Estimators}
\begin{figure*}[!t]
    \addtolength{\belowcaptionskip}{-0.2cm}
    \centering
    \includegraphics[width=1\linewidth, trim={0 5mm 0 0}, clip]{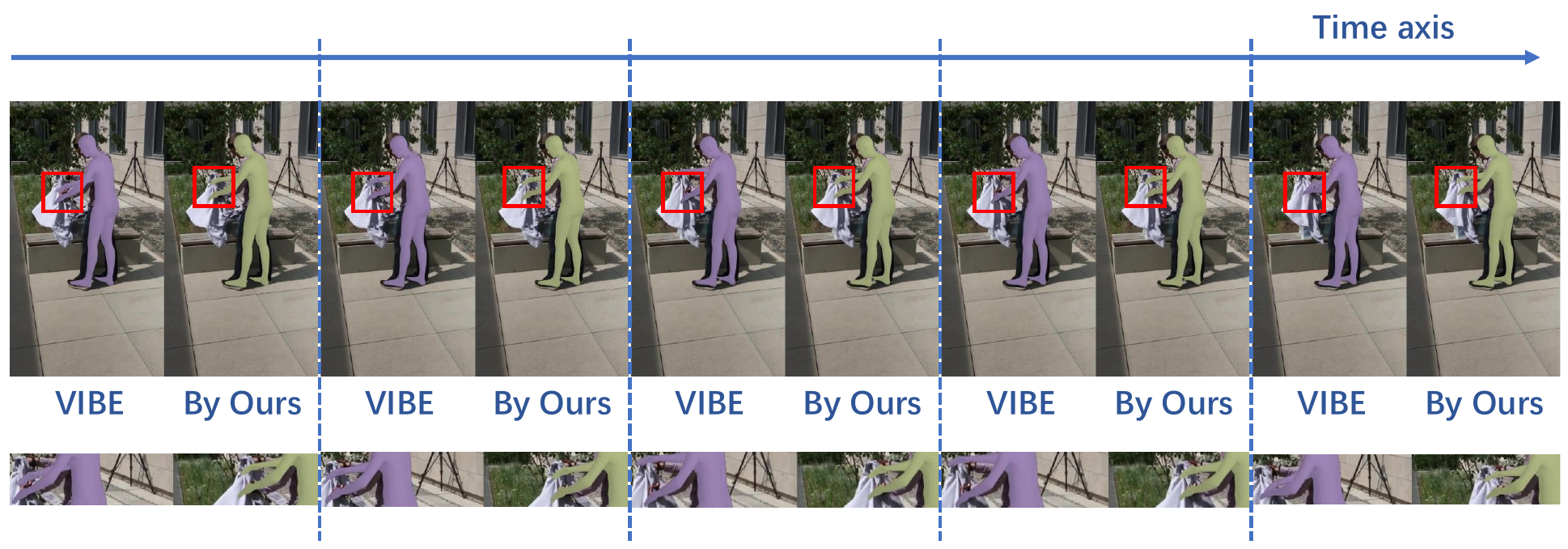}
    \vspace{-0.7cm}
    \caption{
    \textbf{Post-optimization for Human Pose Estimator VIBE.} This figure displays a motion sequence of five consecutive frames. The bottom row shows the enlarged images of the arms. We can see that VIBE produced a sudden jitter of the right arm, while through our optimization, we can mitigate the jitter issues.
    }
    \label{fig:vibe1}
\end{figure*}

\method learns an unsigned distance field of plausible motion and explicitly quantifies human motion dynamics into a score (\ie, distance) which can guide the optimization process. Therefore, our motion prior can be utilized to mitigate jitter issues produced by existing human pose estimators because the motion with jitter movements must be outside the manifold of plausible motion and has a large distance. In Figure~\ref{fig:vibe1}, we present a qualitative comparison with a representative human pose estimator VIBE. For more qualitative results,  please refer to our supplementary video. Besides, as shown in Table~\ref{tab:smooth-only}, the estimation performance often degrades when applying traditional filters (such as one euro) which has been proven in~\cite{zeng2022smoothnet}.
\begin{figure*}[!t]
    \addtolength{\belowcaptionskip}{-1.0em}
    \centering
    \includegraphics[width=1\linewidth, trim={0 0 0 2mm}, clip]{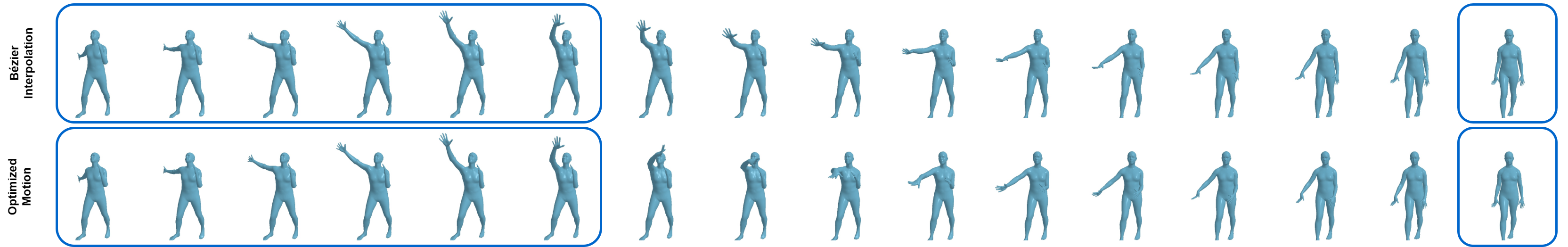}
    \vspace{-0.6cm}
    \caption{
    \textbf{Qualitative Comparison with Bézier Interpolation.} Frames 0 to 5 and frame 15 in the blue boxes are the conditional poses.
    }
    \label{fig:inbetween2}
\end{figure*}


\begin{table}[!t]
\addtolength{\belowcaptionskip}{-0.8em}
\centering
\tiny
\setlength{\extrarowheight}{2pt} 
\begin{tabular}{ccccc}
\hline
\multicolumn{1}{c}{\multirow{2}{*}{\textbf{Method}}} & \multicolumn{4}{c}{\textbf{3DPW}} \\ \cline{2-5} 
\multicolumn{1}{c}{} & \textbf{MPJPE $\downarrow$} & \textbf{PA-MPJPE $\downarrow$} & \textbf{PVE $\downarrow$} & \textbf{Accel $\downarrow$} \\ \hline
SPIN~\cite{kolotouros2019spin} & 99.29 & 61.71 & 113.32 & 34.95 \\
SPIN w/ one euro & 99.53 & 62.24 & 113.55 & 14.23 \\
SPIN w/ S~\cite{zeng2022smoothnet} & 97.81 & 61.19 & 111.5 & \textbf{7.4} \\
SPIN w/ only proposed & \textbf{97.28} & \textbf{60.79} & \textbf{111.4} & 8.55 \\ \hline
EFT~\cite{joo2020eft} & 91.6 & 55.33 & 110.17 & 33.38 \\
EFT w/ one euro & 91.82 & 55.65 & 110.46 & 14.17 \\
EFT w/ S~\cite{zeng2022smoothnet} & 89.57 & 54.40 & \textbf{107.66} & \textbf{7.89} \\
EFT w/ only proposed & \textbf{89.48} & \textbf{53.91} & 107.94 & 9.05 \\ \hline
PARE~\cite{Kocabas_PARE_2021} & 79.93 & 48.74 & 94.07 & 26.45 \\
PARE w/ one euro & 80.46 & 49.32 & 94.81 & 10.52 \\
PARE w/ S~\cite{zeng2022smoothnet} & 78.68 & 48.47 & \textbf{92.5} & \textbf{6.31} \\
PARE w/ only proposed & \textbf{78.61} & \textbf{47.86} & 92.72 & 7.75 \\ \hline
VIBE*~\cite{kocabas2019vibe} & 84.28 & 54.93 & 99.10 & 23.59 \\
VIBE* w/ one euro & 85.89 & 56.49 & 100.80 & 10.87 \\
VIBE* w/ S~\cite{zeng2022smoothnet} & 83.46 & 54.83 & 98.04 & \textbf{7.42} \\
VIBE* w/ only proposed & \textbf{83.14} & \textbf{54.29} & \textbf{97.87} & 8.12 \\ \hline
TCMR*~\cite{choi2021beyond} & 88.47 & 55.70 & 103.22 & 7.13 \\
TCMR* w/ one euro & 90.18 & 57.41 & 104.97 & 6.74 \\
TCMR* w/ S~\cite{zeng2022smoothnet} & 88.69 & 56.61 & 103.40 & \textbf{6.48} \\
TCMR* w/ only proposed & \textbf{88.28} & \textbf{55.69} & \textbf{103.02} & 6.72 \\ \hline
\end{tabular}
\vspace{1em}
\caption{\textbf{Mitigating Jitters on 3DPW Dataset.} "w/ one euro" refers to using the traditional one euro filter for refinement. "w/ S" indicates refinement using SmoothNet. "*" denotes spatio-temporal backbones.}
\label{tab:smooth-only}
\end{table}

\subsection{Motion In-betweening Refinement}
\label{subsec:inbetween}
Moreover, we also evaluate our method with first-order Bézier (linear) interpolation, commonly used in animation software. Specifically, we select frames 0 to 5 and frame 15 as conditional poses which are randomly sampled from AMASS and adopt Bézier interpolation for initial in-betweening, and then we further optimize it with our motion prior. The results are shown in Figure~\ref{fig:inbetween2}. We can see that our method captures human motion dynamics better by guiding the optimization with manifold distances. Bézier interpolation only considers two key frames, while our motion prior takes into account the overall motion trend, so that the right arm still maintains a certain swinging motion before putting it down.

\ssecspace
\subsection{Motion Generation}
\label{subsec:flexible}
\begin{figure*}[!t]
    \addtolength{\belowcaptionskip}{-1.0em}
    \centering
    \includegraphics[width=0.9\linewidth, trim={0 0 0 16mm}, clip]{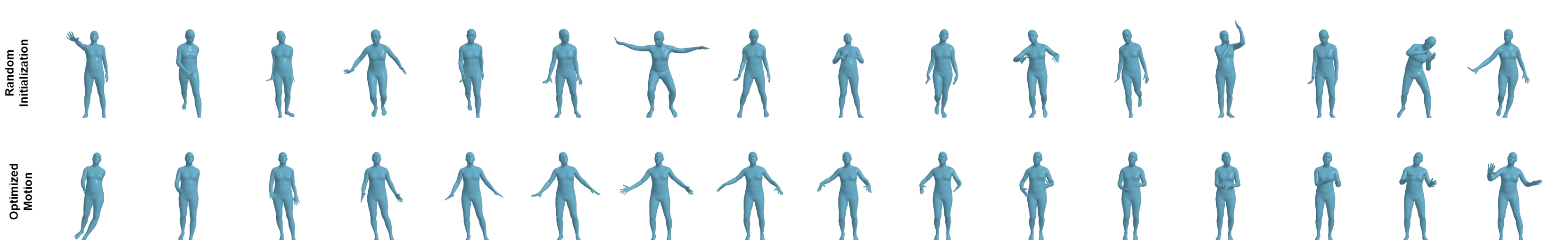}
    \vspace{-0.6em}
    \caption{
    \textbf{Motion Generation.} The first row is the randomly initialized chaotic motion. The second row is the realistic motion we generated, which is the action of closing and subsequently spreading the hands.
    }
    \label{fig:generation}
\end{figure*}

Beyond enhancing the motions produced by existing methods, our approach even has a certain capability of motion generation by converting chaotic sequences into plausible human motions. We begin by randomly selecting 16 varied poses from the AMASS dataset, forming an initial erratic sequence. We then exclusively apply our motion prior, as defined in Eq.~(\textcolor{red}{8}) of the main paper, to this disordered starting point. As illustrated in Figure~\ref{fig:generation}, the generated motion is seamless and natural.

%% file: tables_images_tex/supp_motion_fitting.tex
\begin{figure}
\addtolength{\belowcaptionskip}{-1em}
    \begin{minipage}[t]{0.45\linewidth}
    \begin{center}
    \resizebox{0.9\linewidth}{!}{
    \includegraphics[width=\linewidth, trim={0 0 0 6cm}, clip]{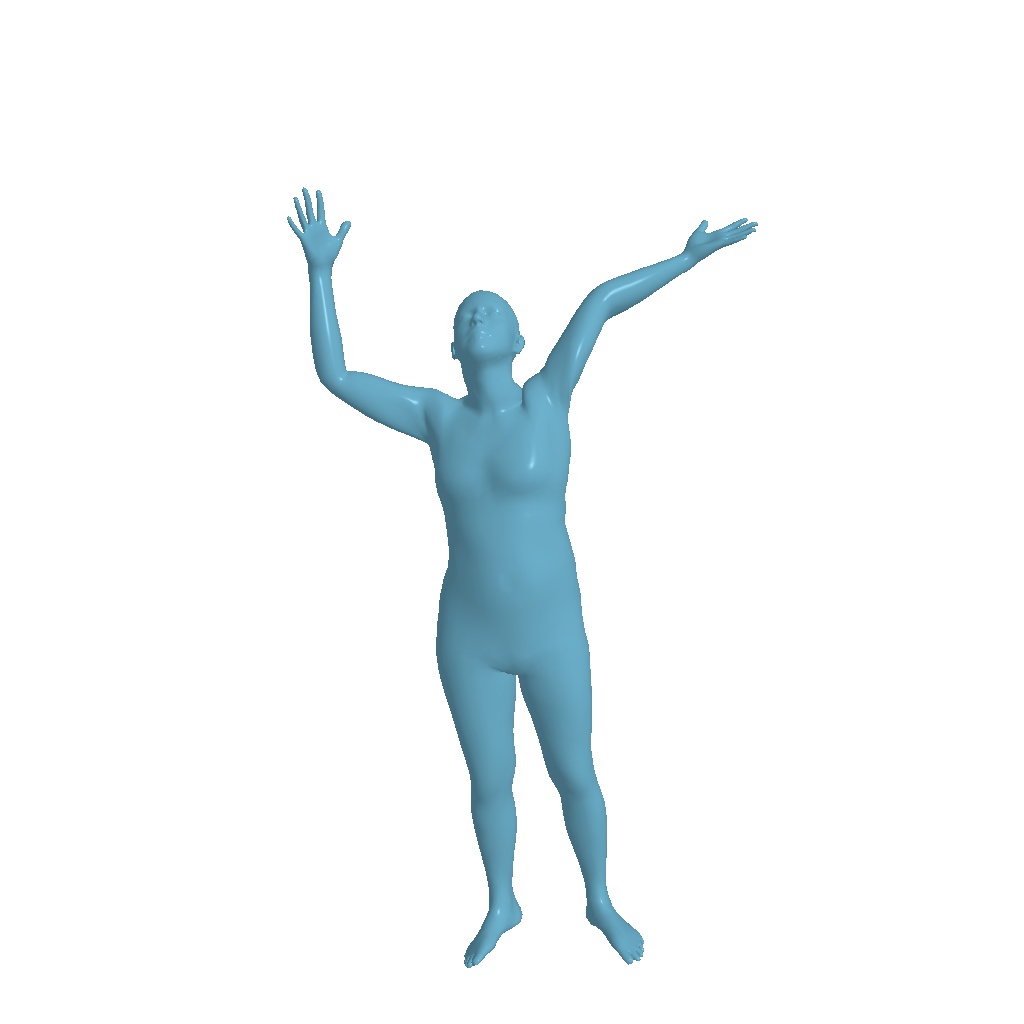}
    }
    \end{center}
    \vspace{-0.43cm}
    \caption{\textbf{HuMoR Accumulation of Errors.}} 
    \label{fig:humor}
    \end{minipage}
    \hfill
    \begin{minipage}[t]{0.52\linewidth}
    \vspace{-12em} %
    \begin{center}
    \resizebox{\linewidth}{!}{
    \includegraphics[width=\linewidth, trim={0 5mm 0 0}, clip]{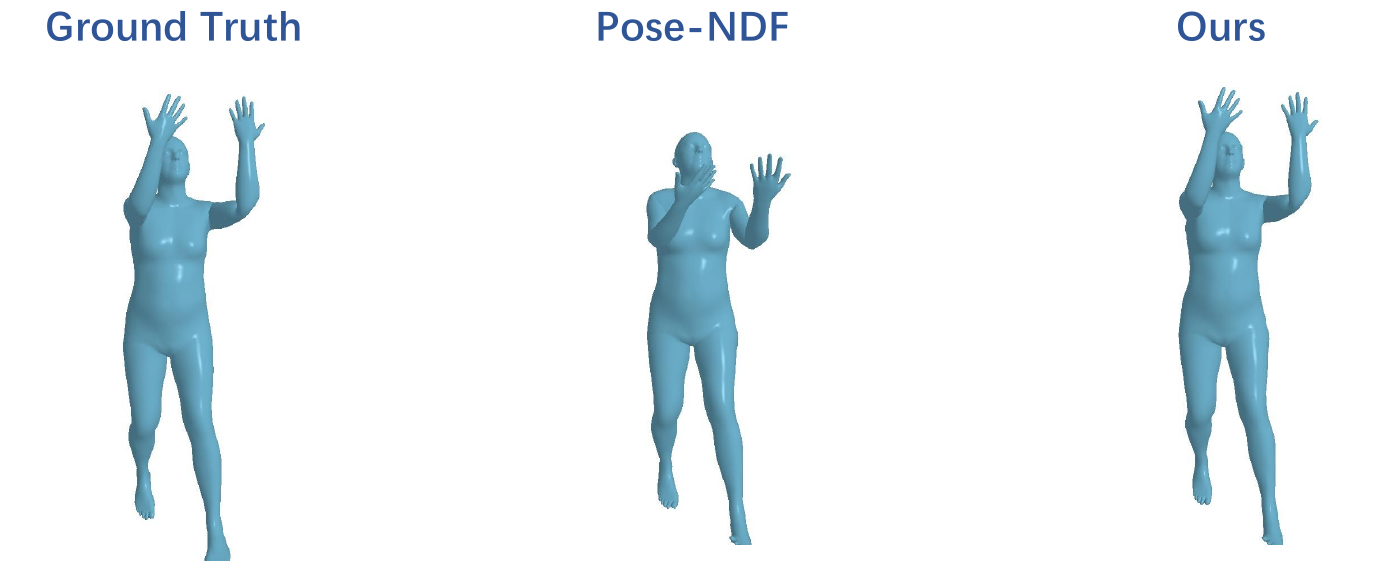}
    }
    \end{center}
    \vspace{-0.5cm}
    \caption{
    \textbf{Motion Range Comparison.} Due to constraints imposed by the traditional temporal regularization term, Pose-NDF struggles to achieve the correct height for arm elevation. In contrast, our method could preserve a more realistic range of motion.
    }
    \label{fig:range}
    \end{minipage}
\end{figure}

%% file: Sections/D_Ablation_Study.tex
\secspace
\section{Ablation Studies}
\label{sec:ablation}
\subsection{Optimal Motion Segment Length}
\label{subsec:seg_length}
\input{tables_images_tex/ablation_length_terms}
\input{tables_images_tex/ablation_fitting}

In this section, we perform the ablation study 
on the experiment of mitigating jitters for human pose estimators. 
We aim to find the optimal motion segment length.
The length of the motion segment $L$ determines the capacity of temporal information. 
Longer motion segments contain more temporal information, but also raise the modeling difficulty and manifold complexity.
We demonstrate the effects on different lengths from 5 to 32 frames in Table \ref{tab:motion_len}. We chose 5 as the minimum motion segment length because the acceleration vector empirically should be at least 3 frames.
Table \ref{tab:motion_len} shows that as the motion segment length increases, all four metrics first decrease and then begin to increase. When the motion segment length $L$ is 16, we can obtain the best performance. 

\ssecspace
\subsection{Impact of Different Temporal Terms}
\label{subsec:terms}
In this section, we explore the influence of different temporal terms on the experiment of mitigating jitters for human pose estimators. 
In the optimization-based tasks, various similar temporal regularization terms (\eg, the sum of joint differences or mesh vertex differences between consecutive frames) are applied to smooth motion. Table \ref{tab:fusion} shows that, naively applying the traditional temporal regularization term Eq.~(\textcolor{red}{7}) to optimize the pose estimator's results can indeed reduce acceleration error and mitigate jitter issues. However, it will lower human pose recognition accuracy, as indicated by MPJPE, PA-MPJPE, and PVE metrics. 
In contrast, by only utilizing Eq.~(\textcolor{red}{8}), our method can not only mitigate jitter issues and smooth motion but also further enhance the pose recognition accuracy. 
Furthermore, we can see that the full optimization function, \ie, an integration of both \method and a traditional temporal regularization term,  will further improve the performance, because it can help jump out of local optima during the optimization process.



\ssecspace
\subsection{Only Utilizing Proposed Prior}
For the experiments of Sec.~\textcolor{red}{4.2}, Sec.~\textcolor{red}{4.3} and Sec.~\textcolor{red}{4.4} in the main paper, we used Eq.~(\textcolor{red}{9}) to regularize motion, which integrates our motion prior with a traditional temporal regularization term. Here, we only use the proposed prior (\ie, Eq.~(\textcolor{red}{8}) in the main paper) in the experiments to demonstrate that even without the integration, we can still outperform the existing SOTAs as shown in Table~\ref{tab:smooth-only}, Table~\ref{tab:denoise-only} and Table~\ref{tab:partial-only}.

For the experiments of Sec.~\textcolor{red}{4.5} and Sec.~\ref{subsec:flexible}, we exclusively apply our motion prior (without the traditional temporal term) as stated in the main paper.

\subsection{Ablation on Losses}
\begin{figure}[!t]
    \addtolength{\belowcaptionskip}{-1.0em}
    \centering
    \includegraphics[width=1\linewidth, trim={0 7.2cm 0 4.3cm}, clip]{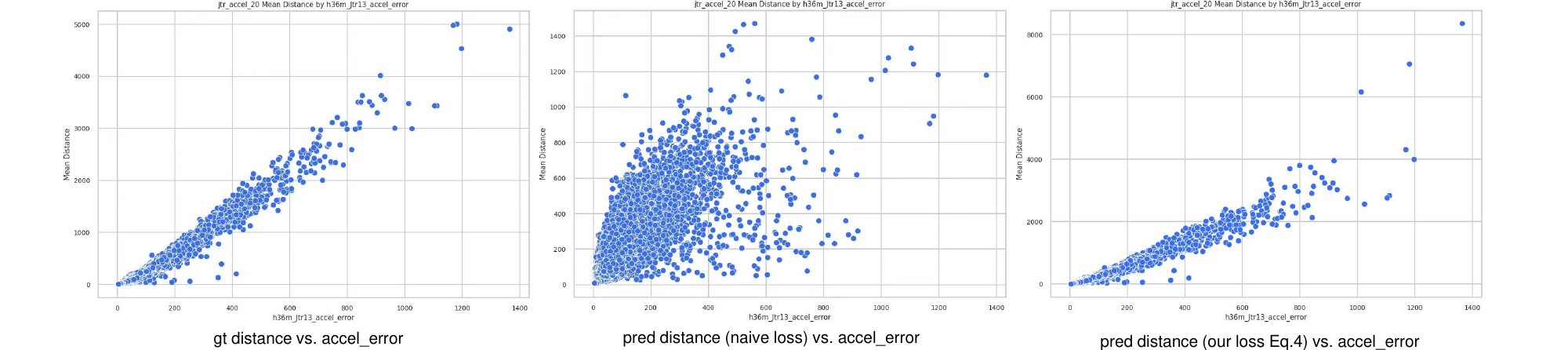}
    \caption{
    \textbf{Ablation on Eq.~(\textcolor{red}{4}).} This is correlation analysis of joint 20, same with Sec.~\textcolor{red}{4.1}. The left one corresponds to gt distances, the middle one corresponds to predicted distances with naive loss, and the right one corresponds to predicted distances with our loss.
    }
    \label{fig:ablation_loss}
\end{figure}
In this section, we conduct the ablation on the loss of Eq.~(\textcolor{red}{4}). In Eq.~(\textcolor{red}{4}), the logarithmic function changes first steeply and then gently, reducing large enough distances to similar values.
This makes the neural network easier to learn, as it will pay more attention to points close to the manifold and will not be affected by points far away. 
In other words, Eq.~(\textcolor{red}{4}) performs non-linear scaling for small and large distances. Figure~\ref{fig:ablation_loss} presents an intuitive comparison, proving that our loss function enables more accurate regression learning (the right one), whereas using a naive loss leads to inaccurate distance predictions (the middle one), thereby making it impossible to reflect the positive correlation with acceleration errors (the left one). For the loss of Eq.~(\textcolor{red}{5}),~\cite{icml2020_2086} has demonstrated it would encourage a 
smoother distance field with unit-norm gradient outside the manifold.

%% file: tables_images_tex/ablation_length_terms.tex
\begin{table}
\addtolength{\belowcaptionskip}{-0.2cm}
    \begin{minipage}[t]{0.48\linewidth}
    \begin{center}
    \resizebox{\linewidth}{!}{
        \scriptsize
        \centering
        \setlength{\tabcolsep}{8pt}
        \setlength{\extrarowheight}{1.2pt} 
        \begin{tabular}{ccccc}
            \hline
            \textbf{Method} & \textbf{\scriptsize MPJPE $\downarrow$ \scriptsize} & \textbf{\scriptsize PA-MPJPE $\downarrow$ \scriptsize} & \textbf{\scriptsize PVE $\downarrow$ \scriptsize}   & \textbf{\scriptsize Accel $\downarrow$ \scriptsize} \\ \hline
            VIBE            & 84.28          & 54.93             & 99.10          & 23.59                \\
            VIBE w/ M-5     & 83.35          & 54.50             & 98.16          & 9.17                 \\
            VIBE w/ M-8     & 83.17          & 54.34             & 97.92          & 8.30                 \\
            VIBE w/ M-16    & \textbf{83.15} & \textbf{54.30}    & \textbf{97.88} & \textbf{8.18}        \\
            VIBE w/ M-32    & 83.32          & 54.48             & 98.11          & 8.44                 \\ \hline
        \end{tabular}
    }
    \end{center}
    \vspace{-0.5em}
    \caption{\textbf{Impact of Motion Segment Length.} We employ \method to optimize the results of VIBE on 3DPW. "M-n" refers to using an n-frame motion segment to model the acceleration manifolds.} 
    \label{tab:motion_len}
    \end{minipage}
    \hfill
    \begin{minipage}[t]{0.48\linewidth}
    \vspace{0.15cm} %
    \begin{center}
    \resizebox{\linewidth}{!}{
    \setlength{\tabcolsep}{8pt}
    \scriptsize
    \centering
    \setlength{\extrarowheight}{1.2pt} 
    \begin{tabular}{ccccc}
        \hline
        \textbf{Method} & \textbf{\scriptsize MPJPE $\downarrow$ \scriptsize} & \textbf{\scriptsize PA-MPJPE $\downarrow$ \scriptsize} & \textbf{\scriptsize PVE $\downarrow$ \scriptsize}   & \textbf{\scriptsize Accel $\downarrow$ \scriptsize} \\ \hline
        VIBE            & 84.28          & 54.93             & 99.10          & 23.59          \\
        VIBE w/ T       & 84.59          & 56.33             & 99.40          & \textbf{7.50}           \\
        VIBE w/ M-16    & 83.15          & 54.30             & 97.88          & 8.18           \\
        VIBE w/ F-16    & \textbf{83.07} & \textbf{54.28}    & \textbf{97.80} & 8.01  \\ \hline   
    \end{tabular}
    }
    \end{center}
    \vspace{-0.3em}
    \caption{\textbf{Impact of Different Temporal Terms.} Through fusion, \method can achieve better performance. "w/ T" denotes with Traditional and "w/ F-16" means that we integrate the traditional term with M-16.
    }
    \label{tab:fusion}
    \end{minipage}
\end{table}

%% file: tables_images_tex/ablation_fitting.tex
\begin{table}
\addtolength{\belowcaptionskip}{-0.5cm}
    \begin{minipage}[t]{0.48\linewidth}
    \vspace{0.2em}
    \begin{center}
    \resizebox{\linewidth}{!}{
        \tiny
        \setlength{\tabcolsep}{8pt}
        \setlength{\extrarowheight}{2pt} 
        \begin{tabular}{ccccccc}
            \hline
            \textbf{Data}       & \multicolumn{2}{c}{\textbf{Noisy HPS}} & \multicolumn{2}{c}{\textbf{Noisy AMASS}} \\ \cline{2-5} 
            \textbf{\# frames}  & 60     & 120     & 60     & 120     \\ \hline{}
            VPoser-t~\cite{SMPL-X:2019} & 3.05 & 4.43 & 5.83 & 6.55 \\
            HuMoR~\cite{rempe2021humor} & 6.08 & 12.67 & 10.28 & 12.63 \\
            Pose-NDF~\cite{tiwari22posendf} & 1.17 & 1.30 & 5.03 & 5.39 \\
                \textbf{Only proposed} & \textbf{0.97} & \textbf{0.98} & \textbf{1.56} & \textbf{1.59} \\ \hline
        \end{tabular}
    }
    \end{center}
    \vspace{-0.2em}
    \caption{\textbf{Motion Denoising}. We compare PVE in cm. "Only proposed" refers to only using our motion prior to regularize the motion without integrating with the traditional temporal term.}
    \label{tab:denoise-only}
    \end{minipage}
    \hfill
    \begin{minipage}[t]{0.48\linewidth}
    \vspace{0.06cm} %
    \begin{center}
    \resizebox{\linewidth}{!}{
    \setlength{\tabcolsep}{8pt}
    \scriptsize
    \centering
    \setlength{\extrarowheight}{3pt} 
    \begin{tabular}{ccccccc}
        \hline
        \multirow{2}{*}{\textbf{Data}} & \multicolumn{2}{c}{\multirow{2}{*}{\textbf{Occ. Leg}}} & \multicolumn{2}{c}{\textbf{Occ. Arm}}            & \multicolumn{2}{c}{\textbf{Occ. Shoulder}}       \\
                              & \multicolumn{2}{c}{}                          & \multicolumn{2}{c}{\textbf{+Hand}}               & \multicolumn{2}{c}{\textbf{+Upper Arm}}          \\ \cline{2-7} 
        \textbf{\# frames}               & 60                    & 120                   & 60            & \multicolumn{1}{c}{120} & 60            & \multicolumn{1}{c}{120} \\ \hline
        VPoser-t~\cite{SMPL-X:2019}      & 8.69         & 10.77         & 8.79           & 10.70           & 8.74                & 10.20                \\
        HuMoR~\cite{rempe2021humor}      & 9.52         & 12.70         & 9.39           & 13.82           & 9.02                & 12.14                \\
        Pose-NDF~\cite{tiwari22posendf}   & 8.50          & 9.40          & 8.66            & 9.43            & 8.73                 & 9.47                 \\
        \textbf{Only proposed}    & \textbf{5.09}         & \textbf{5.33}         & \textbf{5.06} & \textbf{5.26}           & \textbf{5.19} & \textbf{5.32}           \\ \hline
    \end{tabular}
    }
    \end{center}
    \vspace{-0.3em}
    \caption{\textbf{Fitting to Partial Data.} We compare PVE (in cm) on test set of AMASS. Even without the  integration, our results are still better than other methods in all cases.
    }
    \label{tab:partial-only}
    \end{minipage}
\end{table}

%% file: Sections/E_Discussions.tex
\section{Discussions}
\label{sec:discuss}

\subsection{About Joints Decoupling}
 At first, we tried to treat the human body as a whole and used various architectures, including transformers, to model the manifold, but it is hard to learn to map such high-dimensional input to a continuous distance value since extremely large data is required, which is impractical. 
Therefore, we proposed to decouple the joints, reducing the input dimension from 1008 to 42 (taking 16 frames as an example). This makes the data in the low-dimensional space dense enough to capture the data distribution. 

 Despite the decoupling, the joints maintain an inherent correlation through the SMPL model topology and thus reflect human dynamics as a whole. Indeed, this may make it hard to capture the kinematic relationships between joints on different branches, such as left leg and right arm. However, this will not cause pose errors 
when optimizing all joints, since we can always get the correct human body structure via SMPL model.

\subsection{Joints J0-J3 are excluded}
J0 is pelvis, the root joint, which corresponds to the position in the world coordinate system.
J1 is left hip, J2 is right hip and J3 is spine1.
Like previous methods, we set J0 fixed to better capture the changes of human poses in the local coordinate system. 
So J0 is static.
J1-J3 are right next to J0 in the articulated skeleton and therefore have very little movement and very small acceleration, which makes it hard and meaningless to learn distance mapping. 

\clearpage